\documentclass[10pt,journal,compsoc]{IEEEtran}
\ifCLASSOPTIONcompsoc
  \usepackage[nocompress]{cite}
\else
  \usepackage{cite}
\fi
\ifCLASSINFOpdf
\else
\fi
\usepackage{amssymb}
\usepackage{amsmath}
\usepackage{bm}
\usepackage{caption}
\usepackage{graphicx, subfig}
\usepackage{multirow}
\usepackage{soul,color}
\definecolor{Red}{cmyk}{0,1,1,0}
\definecolor{Green}{cmyk}{1,0,1,0}
\definecolor{Cyan}{cmyk}{1,0,0,0}
\definecolor{Purple}{cmyk}{0.45,0.86,0,0}
\definecolor{Rosolic}{cmyk}{0.00,1.00,0.50,0}
\definecolor{Blue}{cmyk}{1.00,1.00,0.00,0}
\definecolor{Orange}{cmyk}{0,0.52,0.80,0}
\definecolor{Black}{cmyk}{1,0,0,1}

\usepackage{url}
\usepackage{hyperref}
\def\eg{e.g.,~}               % for example
\def\ie{\textit{i.e.,~}}      % that is, in other words
\def\sysname{StructNeRF}

\begin{document}
%
% paper title
% Titles are generally capitalized except for words such as a, an, and, as,
% at, but, by, for, in, nor, of, on, or, the, to and up, which are usually
% not capitalized unless they are the first or last word of the title.
% Linebreaks \\ can be used within to get better formatting as desired.
% Do not put math or special symbols in the title.
\title{\sysname{}: Neural Radiance Fields for Indoor Scenes with Structural Hints}
%
%
% author names and IEEE memberships
% note positions of commas and nonbreaking spaces ( ~ ) LaTeX will not break
% a structure at a ~ so this keeps an author's name from being broken across
% two lines.
% use \thanks{} to gain access to the first footnote area
% a separate \thanks must be used for each paragraph as LaTeX2e's \thanks
% was not built to handle multiple paragraphs
%
%
%\IEEEcompsocitemizethanks is a special \thanks that produces the bulleted
% lists the Computer Society journals use for "first footnote" author
% affiliations. Use \IEEEcompsocthanksitem which works much like \item
% for each affiliation group. When not in compsoc mode,
% \IEEEcompsocitemizethanks becomes like \thanks and
% \IEEEcompsocthanksitem becomes a line break with idention. This
% facilitates dual compilation, although admittedly the differences in the
% desired content of \author between the different types of papers makes a
% one-size-fits-all approach a daunting prospect. For instance, compsoc 
% journal papers have the author affiliations above the "Manuscript
% received ..."  text while in non-compsoc journals this is reversed. Sigh.

\author{Zheng~Chen, Chen~Wang,~\IEEEmembership{Student Member,~IEEE}, Yuan-Chen Guo,
        Song-Hai~Zhang,~\IEEEmembership{Member,~IEEE,}
% <-this % stops a space
\IEEEcompsocitemizethanks{\IEEEcompsocthanksitem Z. Chen , C. Wang, Y. Guo and S. Zhang are with the BNRist, Department
of Computer Science and Technology, Tsinghua University, Beijing,
China, 100084.\protect\\
% note need leading \protect in front of \\ to get a newline within \thanks as
% \\ is fragile and will error, could use \hfil\break instead.
E-mail: chenz20@mails.tsinghua.edu.cn, cw.chenwang@outlook.com, guoyc19@mails.tsinghua.edu.cn, shz@tsinghua.edu.cn}% <-this % stops an unwanted space
\thanks{Manuscript received April 19, 2005; revised August 26, 2015.}}

% note the % following the last \IEEEmembership and also \thanks - 
% these prevent an unwanted space from occurring between the last author name
% and the end of the author line. i.e., if you had this:
% 
% \author{....lastname \thanks{...} \thanks{...} }
%                     ^------------^------------^----Do not want these spaces!
%
% a space would be appended to the last name and could cause every name on that
% line to be shifted left slightly. This is one of those "LaTeX things". For
% instance, "\textbf{A} \textbf{B}" will typeset as "A B" not "AB". To get
% "AB" then you have to do: "\textbf{A}\textbf{B}"
% \thanks is no different in this regard, so shield the last } of each \thanks
% that ends a line with a % and do not let a space in before the next \thanks.
% Spaces after \IEEEmembership other than the last one are OK (and needed) as
% you are supposed to have spaces between the names. For what it is worth,
% this is a minor point as most people would not even notice if the said evil
% space somehow managed to creep in.

% The paper headers
\markboth{Journal of \LaTeX\ Class Files,~Vol.~14, No.~8, August~2015}%
{Shell \MakeLowercase{\textit{et al.}}: Bare Demo of IEEEtran.cls for Computer Society Journals}
% The only time the second header will appear is for the odd numbered pages
% after the title page when using the twoside option.
% 
% *** Note that you probably will NOT want to include the author's ***
% *** name in the headers of peer review papers.                   ***
% You can use \ifCLASSOPTIONpeerreview for conditional compilation here if
% you desire.

% The publisher's ID mark at the bottom of the page is less important with
% Computer Society journal papers as those publications place the marks
% outside of the main text columns and, therefore, unlike regular IEEE
% journals, the available text space is not reduced by their presence.
% If you want to put a publisher's ID mark on the page you can do it like
% this:
%\IEEEpubid{0000--0000/00\$00.00~\copyright~2015 IEEE}
% or like this to get the Computer Society new two part style.
%\IEEEpubid{\makebox[\columnwidth]{\hfill 0000--0000/00/\$00.00~\copyright~2015 IEEE}%
%\hspace{\columnsep}\makebox[\columnwidth]{Published by the IEEE Computer Society\hfill}}
% Remember, if you use this you must call \IEEEpubidadjcol in the second
% column for its text to clear the IEEEpubid mark (Computer Society jorunal
% papers don't need this extra clearance.)

% use for special paper notices
%\IEEEspecialpapernotice{(Invited Paper)}

% for Computer Society papers, we must declare the abstract and index terms
% PRIOR to the title within the \IEEEtitleabstractindextext IEEEtran
% command as these need to go into the title area created by \maketitle.
% As a general rule, do not put math, special symbols or citations
% in the abstract or keywords

\IEEEtitleabstractindextext{%

% 目的和背景:本工作的任务是实行少视角下室内场景的新视角合成。在少视角下，NeRF的几何极度欠约束，以至于合成的效果也很差。

%方法：本文吸收了自监督深度估计方法成功的经验，采用了基于patch的多视角一致性photometric loss，约束有纹理区域几何，同时借助平面正则化的方法约束少纹理的几何。通过稠密的自监督深度约束的方法，我们的方法改善了NeRF的几何，同时改善了其视角合成的效果。

%结果和结论：我们在ScanNet, NYUv2, SUN3D室内场景数据集下做了定量的实验，我们的方法展示了其在多个数据集上的优越性 in comparison to the-state-of-art sparse view methods (DSNeRF, Dense Depth Priors)

% \let\oldtwocolumn\twocolumn
% \renewcommand\twocolumn[1][]{%
%     \oldtwocolumn[{#1}{
%     \begin{center}
%           \includegraphics[width=\linewidth]{fig/teaser.png}
%           \captionof{figure}{While solid paints (left top).}
%           \label{fig:teaser}
%     \end{center}
%     }]
% }
%  (See the red arrows that point to the regions with poor synthesis quality \wc{maybe zoom these pointed regions and put them at the bottom right?}).
\begin{abstract}
Neural Radiance Fields (NeRF) achieve photo-realistic view synthesis with densely captured input images. However, the geometry of NeRF is extremely under-constrained given sparse views, resulting in significant degradation of novel view synthesis quality. Inspired by self-supervised depth estimation methods, we propose \sysname{}, a solution to novel view synthesis for indoor scenes with sparse inputs. \sysname{} leverages the structural hints naturally embedded in multi-view inputs to handle the unconstrained geometry issue in NeRF. Specifically, it tackles the texture and non-texture regions respectively: a patch-based multi-view consistent photometric loss is proposed to constrain the geometry of textured regions; for non-textured ones, we explicitly restrict them to be 3D consistent planes. Through the dense self-supervised depth constraints, our method improves both the geometry and the view synthesis performance of NeRF without any additional training on external data. Extensive experiments on several real-world datasets demonstrate that \sysname{} surpasses state-of-the-art methods for indoor scenes with sparse inputs both quantitatively and qualitatively.
\end{abstract}
%  \gycc{``indoor sparse novel view synthesis" is not a word. Define a new word before using it (or simply rephrase). }
% Note that keywords are not normally used for peerreview papers.
}

% make the title area
\maketitle

% To allow for easy dual compilation without having to reenter the
% abstract/keywords data, the \IEEEtitleabstractindextext text will
% not be used in maketitle, but will appear (i.e., to be "transported")
% here as \IEEEdisplaynontitleabstractindextext when the compsoc 
% or transmag modes are not selected <OR> if conference mode is selected 
% - because all conference papers position the abstract like regular
% papers do.
\IEEEdisplaynontitleabstractindextext
% \IEEEdisplaynontitleabstractindextext has no effect when using
% compsoc or transmag under a non-conference mode.

% For peer review papers, you can put extra information on the cover
% page as needed:
% \ifCLASSOPTIONpeerreview
% \begin{center} \bfseries EDICS Category: 3-BBND \end{center}
% \fi
%
% For peerreview papers, this IEEEtran command inserts a page break and
% creates the second title. It will be ignored for other modes.
\IEEEpeerreviewmaketitle

\IEEEraisesectionheading{\section{Introduction}\label{sec:introduction}}
\begin{figure*}[!t]
    \includegraphics[width=\linewidth]{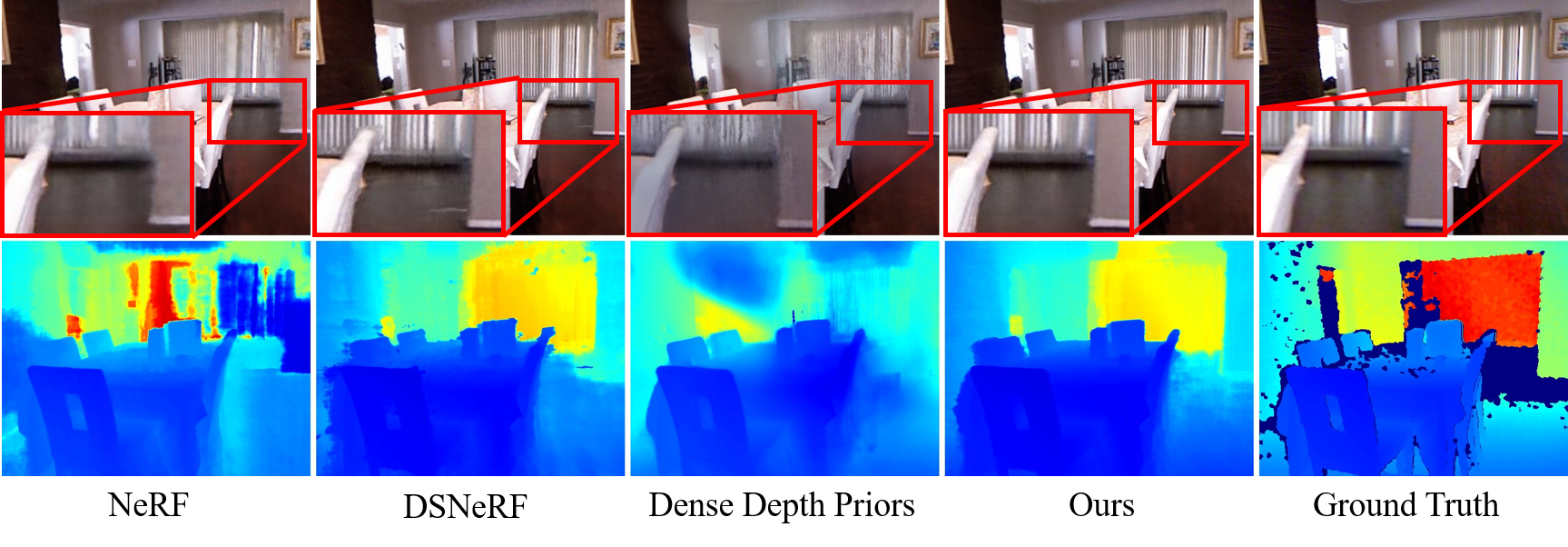}
    \captionof{figure}{Qualitative comparison of novel view synthesis and depth estimation given only a sparse set of indoor images. \sysname{} demonstrates superior performance over state-of-the-arts. We propose two structural hints: patch-based multi-view consistency at textured regions and planar consistency at non-textured regions, significantly improving the synthesis and geometry of radiance fields without any additional data or networks.   } 
    \label{fig:teaser}
% \end{center}
\end{figure*}

% \gycc{The value scale of these depth visualizations should be consistent to enable direct comparisons. Also, it's more common to put the texts (NeRF, DSNeRF ...) below the figure, not above.}
% \gycc{This paragraph is too long. Just need to (1) mention your results are the most satisfying among all alternatives and guide the readers to pay attention to specific regions in the images; (2) briefly introduce what you do to achieve this.}
% NeRF~\cite{nerf} produces unsatisfactory synthesis results due to inaccurate geometry prediction. DSNeRF~\cite{dsnerf} uses insufficient sparse point clouds to constrain the depth of NeRF. While Dense Depth Priors~\cite{dense_depth_priors} enforces dense depth constraints on NeRF's geometry with a pretrained depth completion network, it suffers from both generalization issues and views inconsistency between depth completions.
% \gycc{PAMI papers do NOT seem to have teasers and the abstract directly follows the author list. Consider moving this teaser image into the main text.}

\IEEEPARstart{N}{ovel} view synthesis (NVS) for indoor scenes plays an important role in VR and AR applications, such as virtual navigation through buildings, tourist sites, and game environments. However, people often have to devote extensive efforts to collecting and processing large amounts of input data in order to produce satisfying results~\cite{inside_out,nerf,dense_depth_priors}. It remains to be a problem how to synthesize photo-realistic novel views given limited indoor images~\cite{dense_depth_priors,dsnerf,pixelnerf,regnerf, infonerf}.
%  \gycc{potential reference here}
% \gycc{potential reference here}
%2.NeRF用多层感知机MLP表示一个场景的辐射场和密度分布，并使用体渲染的方式合成各个视角的图像。这个方法表示一些具有复杂几何和外观的场景可以达到非常逼真的效果。然而，如果将NeRF应用到给定几十张图像的真实的室内场景里往往不会达到很好的效果。首先：1.给定稀疏的输入视角，重建物体的几何和外观是不适定的问题，也就是说，NeRF能够很好渲染出训练图像，即使学习了一个不正确的几何形状~\cite{dsnerf}，但是这样的模型会在测试集上产生非常差的渲染结果。室内场景自己的特性更加使得不确定性的问题更加复杂：和对一个中间物体拍照的outside-in的视角场景不同，室内场景的拍摄视角通常是inside-out的视角场景。这样即使是相同的数目的图像，对于场景来说，相邻视角的重叠更加少了，视角相对而言也就显得更加稀疏~\cite{dense_depth_priors}。室内场景的第二个特性是室内通常有很多少纹理的区域，比如墙壁、地面、桌子、天花板等。
Recently, Neural Radiance Fields (NeRF)~\cite{nerf} emerges as a promising technique for NVS. NeRF uses a continuous multi-layer perceptron (MLP) to encode the radiance and density of a 3D scene and then synthesizes novel views through differentiable volumetric rendering. It achieves photo-realistic results even when representing some scenes with complicated geometry and appearance. Nevertheless, sparse indoor scene inputs bring several innate challenges to NeRF. Firstly, reconstructing the geometry and appearance of objects or scenes becomes an ill-posed problem with insufficient inputs. Even though NeRF can well fit the training images at the pixel level, the geometry is indeed inaccurate and leads to unsatisfying renderings at test viewpoints~\cite{dsnerf}. The necessity of "inside-out" view capture for indoor scene images exaggerates this issue~\cite{hedman2016scalable}. Compared with "outside-in" viewing scenarios for outdoor scenes or standalone objects, adjacent views would have less overlap with each other given the same number of images~\cite{dense_depth_priors}. Secondly, indoor scenes contain many textureless regions such as walls, floors, tables, and ceilings, making it hard for NeRF to find enough cross-view 3D correspondences.

% 最近一些工作开始尝试利用深度先验去解决NeRF在室内新视角合成，比如DSNeRF采用了SFM得到的稀疏深度约束NeRF的深度，但是这个深度是非常sparse，无法约束大部分的像素区域，尤其是少纹理区域；DenseDepthPriors利用一个预训练好的深度补全网络来估计各个视角的深度，利用得到深度结果去约束NeRF的深度。但是这个深度补全网络不是视角一致的、而且泛化性并不好。本文借鉴最近几年的自监督的深度估计方法的ideas~\cite{}，采取了基于patch的多视角一致的photo-metric loss去约束NeRF各个视角每个像素的深度，我们的自监督深度约束是稠密的、多视角一致的,而且不需要深度补全网络，也就避免了泛化困难问题。
% 对于少纹理区域，我们利用训练视角提取出超像素，将每个超像素近似为平面，对于每个大面积的平面区域，我们采用了共平面的深度假设~\cite{plnet}，具体而言我们使用平面一致性loss来约束NERF少纹理区域的深度。

Several recent studies~\cite{dsnerf, nerfingmvs, dense_depth_priors} leverage depth priors to improve the performance of NeRF in novel view synthesis. DSNeRF~\cite{dsnerf} adopts the sparse depth point cloud from COLMAP~\cite{colmap1} to directly constrain the depth rendered by NeRF. However, the depth from Structure-from-Motion (SfM) is both sparse and noisy. Dense Depth Priors~\cite{dense_depth_priors} further utilizes a depth completion network to predict dense depth maps, which are then used to guide the sampling and depth prediction of NeRF. However, the depth completion network introduces view inconsistency and generalization issues. To overcome these problems, we present \sysname{}, a technique that takes inspiration from recent self-supervised depth estimation methods~\cite{p2net, plnet} and incorporates structural hints naturally contained in multi-view inputs, which turns into easy-to-adapt regularizations for NeRF geometry without any additional networks or data. \sysname{} considers the huge differences between textured and textureless regions and tackles them separately. Inspired by the insight of NeRF++~\cite{nerf++}, we notice that the ability of NeRF to model the appearance of view-dependent effects leads to the ambiguity between its 3D shape and radiance (shape-radiance ambiguity). To reduce this ambiguity, we ensure that the same 3D region in different views is view-consistent by leveraging a patch-based multi-view consistent photometric loss based on depth warping. The resulting depth constraints are therefore dense and view-consistent. Patch-based photometric loss works well for textured regions, but it fails to discriminate non-textured regions that are common in indoor scenes, such as floor, walls, tables and ceiling. At the same time, we notice these regions are almost planes, so we further restrict them to be planar. To be more specific, we segment each input view into superpixels and group them as plane priors (Most superpixels are planes as shown in Fig.~\ref{fig:superpixels}). Then the co-planar constraint~\cite{plnet} is applied to constrain the depth of those regions. Additionally, we apply warm-up training to reduce the negative impact of noisy sparse point clouds from COLMAP, which helps \sysname{} to better utilize the sparse depth information. We evaluate our method on three indoor datasets: ScanNet~\cite{scannet}, NYUv2~\cite{nyuv2}, SUN3D~\cite{sun3d}, using only sparse inputs. Results show that our proposed structural hints and training strategy together enable \sysname{} outperform existing per-scene training methods (only use data of target scene without additional training). Compared to methods that utilizes external data to train their depth estimation networks (we call data-driven methods), our method still shows comparable performance on their pretrained datasets and surpass them on other ones. A demonstration of the comparisons can be found in Fig.~\ref{fig:teaser}. 
% \gycc{The word ``pretrain" comes from nowhere and will confuse the readers. Explain or rephrase.}\wc{We have added explanation for these words.}

% 总而言之，我们的方法改进了少视角下的室内场景新视角合成，基于以下贡献：
% 1.基于patch-based 多视角一致性 photometric loss，我们的NeRF获取了视角一致的、稠密的自监督深度约束,同时避免了泛化困难的问题。
% 2.我们借助超像素来近似平面，结合平面深度正则化的方式，使得NeRF在少纹理区域也能得到合理的几何质量，这样进一步提高了视角合成的质量。
% 3.在使用了少数的输入图像的情况下，我们在室内场景数据集ScanNet和NYU数据集上评估了我们的方法，我们展示了我们的方法相对于最近在NeRF上使用了基于SparseDepth或者multi-view stereo (MVS) 方法的工作基础上改进了。而相对于~\cite{dense_depth_priors}，我们的方法不需要预训练的深度补全网络的方法，也就没有泛化困难的问题。
% 4.看结果：并且通过实验证明，得益于鲁棒的多视角一致性和少纹理区域的平面正则化，我们的方法在一定程度上还能够改进DenseDepthPriors的效果。
The contributions of our paper can be summarized as the following:

1) By leveraging the patch-based multi-view consistent photometric loss, \sysname{} obtains dense and view-consistent depth constraints, without the need for depth completion network trained on external data.

2) We re-project points in textureless regions into the 3D space and enforce them to be planes with the plane consistency loss. Therefore, the reconstructed planes are more flat and the rendering quality is also improved.

% \gycc{Too brief. What is plane consistency loss? Also, ``more flat planes" is ambiguous: does it mean the planes are more flat or the there are more number of flat planes?}\wc{edited}

3) With the simple yet effective warm-up training strategy minimizing the noises coming from SfM reconstructed point clouds, we achieve state-of-the-art novel view synthesis performance given limited input views. 

\begin{figure*}[h]
\centering
\includegraphics[width=1.0\linewidth]{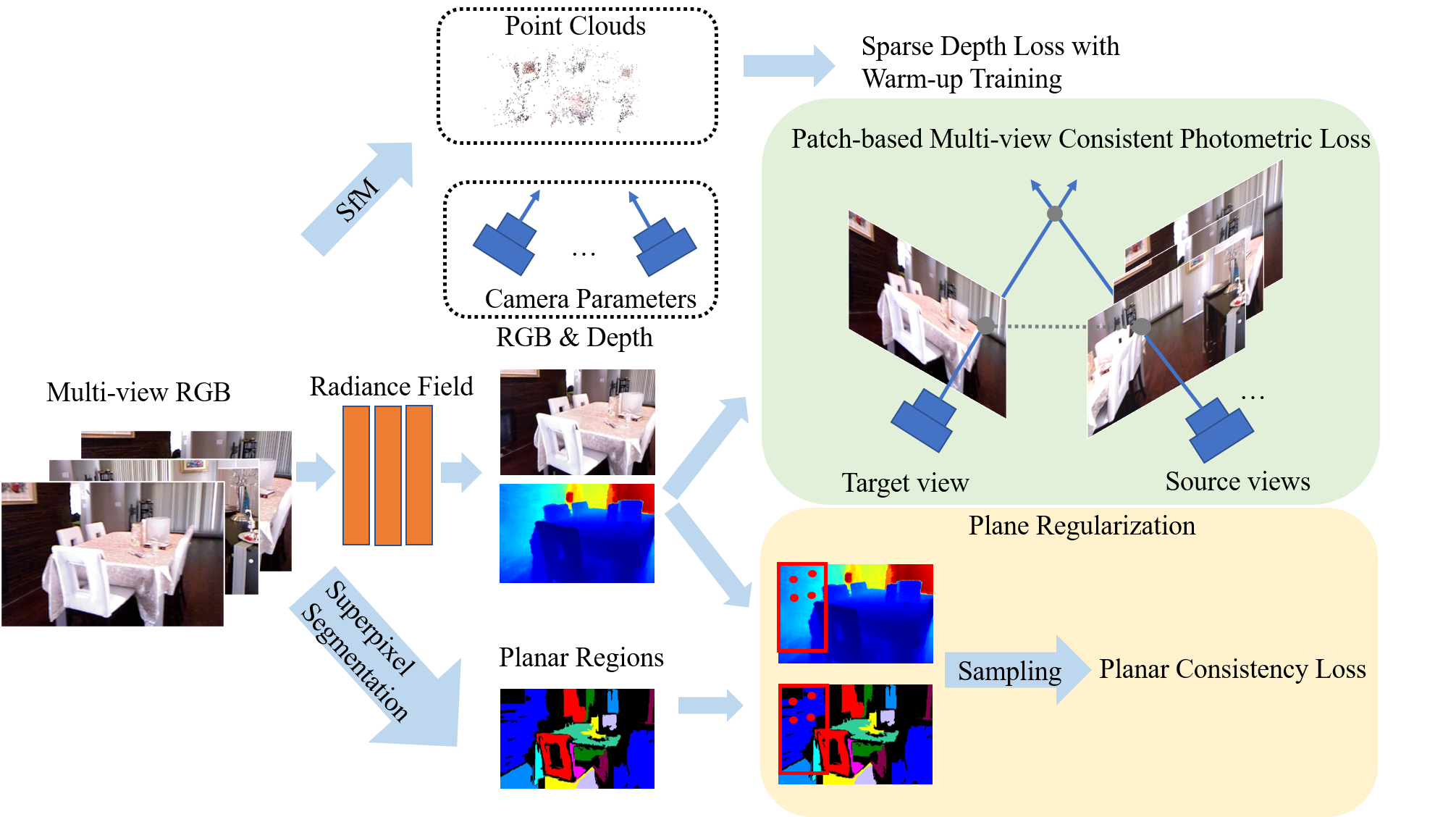}
\caption{We exploit the inherent structural hints in sparse views to improve the performance of NeRF in novel view synthesis. Firstly we utilize the structure-from-motion to obtain camera parameters and sparse point clouds. The sparse point clouds are used to constrain the depth of NeRF at keypoints featuring a warm-up training scheme. For regions with rich textures, we utilize the patch-based photometric loss for self-contained dense depth supervision. We also propose a planar consistency loss to regularize the depth of non-texture regions with the assistance of superpixel segmentation.}
\label{fig:pipeline}
\end{figure*}

\section{Related Work}
In this section, we briefly review Neural Radiance Field with sparse inputs and self-supervised depth estimation.

%1.介绍基于显式方式的渲染和它们表达上的局限性
%2.NeRF
%3.引入少视角的问题：几何很差，基于深度优化的方法和它们的局限性
%4.引入我们的方法，相对于他们的优势

% 1.MPI:Stereo magnification: Learning view synthesis using multiplane images
% 2.mesh:3d photography using context-aware layered depth inpainting
% 3.Point clouds:SynSin: End-to-end view synthesis from a single image
%尽管这些方法能够在测试时能很快地渲染出来，但是他们的表达性有限的。

\subsection{Neural Radiance Fields with Sparse Inputs} 
% \gycc{``Neural Radiance Fields with Sparse Inputs" would be better. ``Novel View Synthesis" is too big and you don't really spend much words on methods other than NeRF (also they are not quite related to this paper).}
% Novel view synthesis is a long-standing problem in the computer graphics and computer vision community. Traditional image-based rendering approaches mainly follow the procedure of warping and blending input views to target ones~\cite{gortler1996lumigraph, levoy1996light}. Geometry representations for view synthesis include explicit representations such as multiplane images~\cite{mpi, mildenhall2019local, srinivasan2019pushing}, voxel grids~\cite{henzler2020learning, kalantari2016learning, penner2017soft}, mesh~\cite{3dphoto} and point clouds~\cite{synsin}. Nevertheless, these explicit representations are discrete and often require abundant data and memory. The rendering quality is also limited by the accuracy of the reconstructed geometry. \gycc{This paragraph would be unnecessary if you decide to change the subsection title.}

%NeRF通过连续的体辐射场的隐式表示方式避免了体素方式的陡峭。。的属性，通过体渲染的方式优化每个视角的颜色重建loss，达到了Photo-realistic 的效果。
Based on implicit neural representations, Neural Radiance Fields (NeRF)~\cite{nerf} encoded 3D scenes into a continuous multi-layer perceptron (MLP) and achieved photorealistic novel view synthesis. A growing number of NeRF extensions then emerged, e.g., reconstructing without camera poses~\cite{lin2021barf, wang2021nerf--}, modelling non-rigid scenes~\cite{park2021nerfies, park2021hypernerf}, unbounded scenes~\cite{zhang2020nerf++}, handling reflections~\cite{guo2022nerfren, verbin2021ref} and super-resolution~\cite{wang2021nerf}. When the scene is observed by sparse views, NeRF would 
however estimate a wrong density distribution, which is specifically reflected as some artifacts in the rendering process, such as "floaters". Here we give a detailed review of NeRF-based methods in both object-level and scene-level when the inputs are sparse.
%然而当场景的观察视角稀疏时，NeRF建模几何和外观的高表达能力，反而会导致一个错误的密度分布，其结果就是在渲染过程中的一些瑕疵，比如"floaters"。

Given sparse object-level views, several recent works~\cite{srf,mvsnerf,dietnerf,codenerf,pixelnerf} synthesized novel views using a pretraining with an optional per-scene optimization strategy. The pretrained network is however not suitable for indoor scenes due to the domain gap. Other methods impose regularizations on NeRF geometry, for example, RegNeRF~\cite{regnerf} samples unobserved camera poses and regularizes patches rendered from those views with a depth smoothness loss and a trained normalizing flow model respectively. InfoNeRF~\cite{infonerf} utilizes regularization based on information theory to improve view synthesis. However, both InfoNeRF~\cite{regnerf} and RegNeRF~\cite{regnerf} do not guarantee multi-view consistency. And all these object-level approaches never take into account the characteristics of the indoor scenes that we mentioned in Sec.\ref{sec:introduction}.

% NerfingMVS,DSNeRF, DenseDepthPriors
%NerfingMVS使用MVS得到的深度来过拟合一个对这个场景的深度预测器，并使用得到的深度先验指导NeRF的采样过程。然而他们的方法缺少对NeRF的深度进行直接的约束。
% DSNeRF直接利用COLMAP预测得到的稀疏深度信息对NeRF的深度进行约束，为了解决稀疏重建过程中的不准确性，所有3D点都根据重投影误差被进行了加权。虽然得到的稀疏点云是天然的多视角一致的，但是它只能对极少部分的像素进行深度约束。
% DenseDepthPriors将COLMAP得到的稀疏点云投影到各个视角，并通过一个预训练的深度补全网络得到每个视角的稠密深度图，利用这个深度稠密图来对NeRF的深度进行直接监督，并借鉴NerfingMVS的过程利用这个深度先验对NeRF的采样过程进行指导。然而这个深度补全网络泛化性是在其他数据集的泛化性有限，并且这个深度补全过程不是多视角一致的。
With regard to scene-level, recent studies like NerfingMVS~\cite{nerfingmvs}, DSNeRF~\cite{dsnerf} and Dense Depth Priors~\cite{dense_depth_priors} proposed to introduce depth priors to resolve the unconstrained geometry problem in NeRF from different aspects. Without additional network or training, DSNeRF~\cite{dsnerf} utilizes the sparse depth information from COLMAP~\cite{colmap1} directly to constrain the depth rendered. NerfingMVS~\cite{nerfingmvs} instead trains a monocular depth estimation network to get scene-specific depth priors for guiding NeRF sampling. Similarly, Dense Depth Priors~\cite{dense_depth_priors} leverages a pretrained depth completion network to predict dense depth maps for each view individually, which are then used to both supervise the rendered depth and guide NeRF sampling. However, there are two obvious problems in Dense Depth Priors. Firstly, the depth completion network is not view-consistent because each view is processed individually. Secondly, it also suffers from generalization issues as it relies on labeled training data such as ScanNet~\cite{scannet}.

% 相对于DenseDepthPriors，我们的方法不依赖深度补全网络利用基于patch的自监督多视角一致性的深度loss可以直接对NeRF得到的深度进行稠密的监督。同时，我们借助共平面一致性loss方式，在室内少纹理的区域进行了正则化，解决了纹理不足无法多视角match的缺陷。

Compared with previous methods, \sysname{} leverages patch-based multi-view consistent depth loss to obtain dense supervision for NeRF without any depth completion network and additional data (unlike Dense Depth Priors~\cite{dense_depth_priors}). Besides, we are the first to utilize a 3D planar consistency loss to further improve the quality of view synthesis in texture-less regions, which is also complementary to the multi-view patch-match loss.

% \gycc{Find the KEY points for the advantages of your method. For example, (1) the patch-based multi-view consistency works without any external data; (2) this paper proposes the first NeRF-based method to attempt to improve texture-less regions.}
%\cz{edited}

% \subsection{NeRF with sparse inputs}

%part:
\subsection{Self-supervised Depth Estimation}
% self-supervised自监督的深度估计被提出来用以减轻对大批带深度标签训练数据的依赖。
% SfMLearner~cite{sfmlearn}是一个开创性的工作，它通过photometric loss来监督来自PoseNet和DepthNet的几何估计。
Self-supervised depth estimation methods are proposed to ease the demand for large-scale labeled training data. SfMLearner~\cite{sfmlearner} is a pioneering work that supervises the geometry estimations from a depth estimation network by photometric loss. To solve the issue of dynamic objects, optical flow methods are used to compensate for the moving pixels. Semantic masks provided by pretrained semantic segmentation models are also utilized to handle dynamic objects~\cite{signet}. The approaches do not get satisfactory results in indoor scenes because they do not take into account the non-texture regions. %\wc{add reason}.

% 为了解决动态物体问题，在方法文献()里~\cite{}光流被用来补偿移动的像素,并且通过预训练的语义分割模型提供出的语义masks被用来分别处理可能的动态物体。

%比如 SIGNet: Semantic Instance Aided Unsupervised 3D Geometry Perception

% 但是上面的方法没有在室内场景下取得满意的结果，
MovingIndoor~\cite{movingindoor} is the first self-supervised depth estimation approach focusing on indoor scenes. The authors propose to use the sparse flow via matching with SURF~\cite{surf} to initialize the optical flow estimation network, SFNet. In the training process, sparse flows are propagated from textured regions to non-textured regions through iterations and finally transformed into dense flows, which are then used to supervise the depth estimation network. P\({}^{\mbox{2}}\)Net~\cite{p2net} leveraged a patch-based multi-view consistency photometric error to constrain the depths. Other methods also adopt structural regularities such as co-planar constraints to improve depth estimations~\cite{p2net, structdepth, plnet}.

%MovingIndoor方法是首个研究室内场景下的自监督估计方法，他们提出用通过SURF~\cite{}匹配结果得到的稀疏光流进行初始化的光流估计网络SFNet，在训练过程中，稀疏光流通过迭代的方式从有纹理区域传播到无纹理区域，最终转变成稠密的光流。这个得到的稠密光流被用来作为Pose和Depth学习的监督信号。

% P2Net利用基于patch的多视角一致性的loss来约束depth和pose。

% 并且在文献P2Net、StructDepth、PLNet、通过共平面的结构化正则化先验被用来解决室内少纹理的问题。
% Our methods 
Motivated by these self-supervised indoor depth estimation approaches, we propose to utilize the structural hints naturally embedded in indoor scenes to constrain the depth of NeRFs, \ie the patch-based multi-view consistency loss from~\cite{p2net} and planar consistency loss from~\cite{plnet}. However, previous work mainly focus on the depth estimation task, we are the first to introduce these priors in NeRF and demonstrate that they can significantly resolve the unconstrained NeRF geometry issue and enable higher quality view synthesis.

% \wc{(Mention our difference here.) } 
%\cz{edited}

% \gycc{Could be more specific. Motivated from which method? What structral hints? Don't be afraid to repeat what you have said earlier.}
%\cz{edited}

% Although our method adopts similar formulations as~\cite{p2net, plnet}, the indoor scenes are modeled with NeRF for its state-of-the-art performance in novel view synthesis. By leveraging the self-supervised depth priors, our method achieves appealing results for indoor scenes from sparse input views. (王琛：最好不要说我们和他们很像，就说受他们启发就行了）

% 我们的方法参考了他们的思路并用到了NeRF上。我们借助基于patch的表示方法进行了多视角的几何一致性约束，利用共平面约束解决了室内少纹理场景的几何约束困难问题。

%与自监督的深度估计方法不同的是，我们直接在NeRF上面进行了稠密的匹配，去除了稀疏关键点的提取过程，也不需要预训练的网络。我们发现在我们的消融实验里进行稠密的匹配，比只进行稀疏匹配效果要更好。

% 我们的方法吸收了自监督深度估计的方法优点，利用自监督方法改善了NeRF的几何（保证了视角一致性和共平面的正则化），从而优化了视角合成的质量，同时不依赖大数据集的预训练，。

\section{Method}
%Self-Supervised Nerf
%先概述DSNERF和NERF
%在这篇文章里，我们借助两个自监督的深度先验去改进NERF的几何:基于Patchmatch多视角一致性、基于共平面的正则化。

% \begin{figure}
%   \centering
%   \includegraphics[width=1.0\textwidth]{fig/pipeline.png}
%   \caption{Pipeline of our method.}
%   \label{img} 
% \end{figure}
% \setcounter{figure}{1}

%我们的方法推进了少视角的室内新视角合成(shown in Fig.~\ref{fig:pipeline}).首先，我们利用SfM得到了相机参数和稀疏的点云. 我们将自监督的深度估计的方法融入到NeRF的优化中，通过添加patch-based multi-view consistent photometric loss~\ref{sec:patch-match}约束~和共平面一致性loss~\ref{sec:plane-regularization}. Lastly, we adopt a warm-up training strategy to reduce the influence of noisy sparse point clouds when adding the sparse depth priors to NeRF~\ref{sec:training}.

\sysname{} facilitates indoor novel view synthesis given only sparse input images, the framework of which is shown in Fig.~\ref{fig:pipeline}. Firstly, we obtain sparse point clouds and camera parameters from Structure-from-Motion (SfM). We then incorporate self-supervised depth estimation methods into the optimization of NeRF by imposing patch-based multi-view consistent photometric loss (Sec.~\ref{sec:patch-match}) and planar consistency loss (Sec.~\ref{sec:plane-regularization}). Lastly, we observe that while point clouds from SfM could serve as sparse depth priors for NeRF, it suffers from noisy estimation, for which we adopt a warm-up training strategy to gradually decay its contribution to the entire optimization (Sec.~\ref{sec:training}). Before introducing our method, we briefly revisit NeRF~\cite{nerf} in Sec.~\ref{sec:preliminaries}.

\subsection{Preliminaries}
\label{sec:preliminaries}

Neural Radiance Fields (NeRF) represents a scene as a continuous neural volume using a multi layer perceptron (MLP) $f_\theta:(\mathbf{x}, \mathbf{d})\rightarrow (\mathbf{c}, \sigma)$ with $\theta$ as the learnable parameters, where $\mathbf{x}\in\mathbb{R}^3$ and $\mathbf{d}$ denotes a 3D position abd the view direction, $\sigma\in \mathbb{R}$ and $c\in\mathbb{R}^3$ the corresponding density and radiance.

% The MLP
%为了渲染一个像素，MLP首先评估从相机光线上r=o+td采样出的点来获取他们的密度和radiance。然后C(r)通过体渲染公式使用积分求解：
NeRF is an emission-only model, which means the color of a pixel only depends on the radiance along a ray with no other lighting factors. Therefore, according to volume rendering, the color along the camera ray $\mathbf{r}(t) = \mathbf{o}+t\mathbf{d}$ that shots from
the camera center $\mathbf{o}$ in direction $\mathbf{d}$ can be approximated by numerical quadrature:
%Volume rendering:
\begin{equation}
    \hat{\mathbf{C}}(\mathbf{r}) = \int_{t_n}^{t_f}T(t)\sigma(t)\mathbf{c}(t)dt,
\end{equation}
where $T(t)={\rm{exp}}(-\int_{t_n}^{t}\sigma(s)ds)$.

%NeRF通过最小化如下的loss来优化的
NeRF is optimized by sampling random rays from all training images and minimizing the rendered and ground truth pixel color in L2 norm:
\begin{equation}
    L_{\rm{Color}}=\sum \lVert\hat{\mathbf{C}}(\mathbf{r}) - \mathbf{C}(\mathbf{r}) \rVert_2^2
\end{equation}

\subsection{Patch-based Multi-view Consistency}
\label{sec:patch-match}

%motivation:
% NeRF因为具有shape-radiance ambiguity. 
% DSNERF缺少稠密的深度约束，本部分借助自监督的多视角的深度先验来约束每个像素的深度；
% 首先，我们需要获取Nerf每个采样点的深度

%  ambiguity:虽然NeRF具有建模view-dependent appearance的外观能力，但是也导致其在3D形状和radiance之间的模糊性，也会导致一些退化解~\cite{nerf++}

The ability of NeRF to model the appearance of view-dependent appearance leads to the ambiguity between its 3D shape and radiance~\cite{nerf++}. To reduce the shape-radiance ambiguity,
we leverage multi-view consistency explicitly to supervise the depth of every pixel for each view.

To begin with, we render the depth of a given pixel with the formulation proposed in the original NeRF paper. The depth of the ray $\mathbf{r}(t)={\mathbf{o}}+t{\mathbf{d}}$ can thus be calculated as the following:

\begin{equation}
\hat{\mathbf{D}}(\mathbf{r})=\int_{t_n}^{t_f}T(t)\sigma(t)tdt,
\end{equation}
%, where ...

%equation(TODO)

%我们采用基于patch的表示方式
%1.patch-match：为了更好的效果，我们采用了基于patch的方式进行多视角的匹配,也就是我们利用以一个点为中心的周围的窗口表示方式。基于patch的方法相较于基于点的方式对于网络来说学习起来更加具有判别能力~\ref{p2net}

% 需要放一张详细的解释图TODO
% 在介绍基于Patch的方式之前，我们先展示基于点的多视角一致性深度约束是怎么做的：
% 我们从目标视角t采样点$p^t_i$,(注意后面的t和NeRF里定义不一样,NeRF里的t是表示光线上采样区间长度，这里的t表示的是targetview的帧序号，如果不做特殊说明，后面默认t表示目标视角的帧序号) 原先基于点的warp过程将这些点重投影到源视角$I_s$, 利用公式:

\begin{figure}[ht]
\centering
\includegraphics[width=1.0\linewidth]{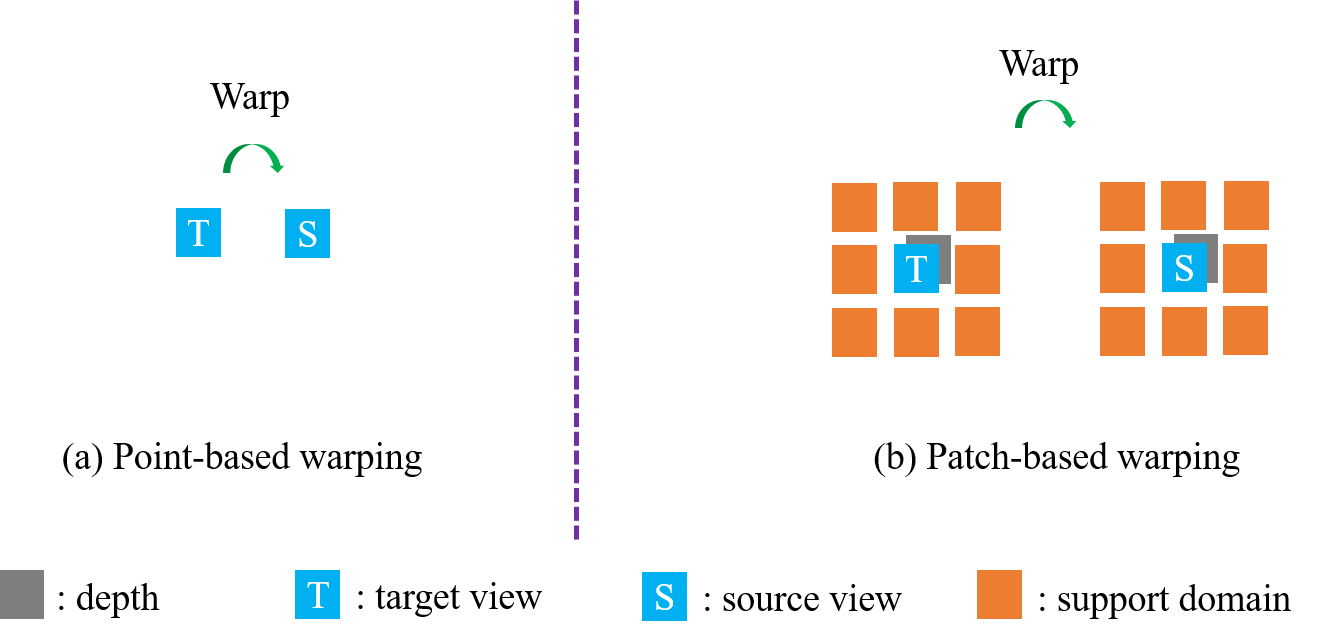}
\caption{The point-based and patch-based warping operations. Point-based operation warps pixel-by-pixel and suffers from severe false matching. However, patch-based operation warps a pixel together with its support domain from the target view to the source view, leading to more robust representations in the depth estimation task. This figure is adopted from \cite{p2net}.}
\label{fig:patchmatch}
\end{figure}

% \wc{This figure is adopted from \cite{p2net}.}
% \cz{}

After we sample points $p^t_i$ from the target view $I_t$, the original point-based warping process back-projects the extracted points to the source views $I_s$ by,
\begin{equation}
p^{t \rightarrow s}_i=KM^{s}{M^{t}}^{-1} (\hat{\mathbf{D}}(p_i)\odot(K^{-1}p^t_i)),
\end{equation}
where $K$ denotes camera intrinsic parameters. $M_s$ and $M_t$ are the camera extrinsic parameters of the source view $I_s$ and the target view $I_t$ respectively. $\hat{\mathbf{D}}(p_i)$ is the rendered depth at the pixel $p_i$. $\odot$ represents Hadamard Product.% \wc{explain $\odot$ operation}.

%然而，基于点的表示方式不够具有判别力，可能导致错误的匹配，因为一张图像里很多像素具有相同的intensity values.
%在传统的SLAM里和自监督的深度估计方法里，为了克服上述问题，一个支持域被定义为点周围的局部窗口。具体 而言Specifically, Photometric loss 在这个支持域里累积计算而不是单个孤立的点，这使得融合了支持域的采样点更加独特。

Nevertheless, point-based representation is not discriminative enough and may cause false matching because many pixels have the same intensity values in an image. Similar to traditional SLAM pipelines~\cite{dso} and self-supervised depth estimation~\cite{p2net}, we define a \textit{support domain} $\Omega_{p_i}$ as the local window surrounding the sampled point $p_i$. Photometric loss is calculated over each supported domain instead of an isolated point, with which the depth of NeRF can be more accurate because the sampled points combined with their support domains are more unique (Fig.~\ref{fig:patchmatch}).

%在支持域里，每个点的深度被假设是相同的，。基于patch的warp过程就是使用相同的深度，通过Nerf渲染出的采样点的深度，将采样点和它的支持域warp到其他视角，使用公式：

In each support domain, the depth of every point is assumed to be the same. We then use the same depth (the depth of the sampled point in the center of each patch), rendered by a neural radiance field, to warp the sampled point together with its support domain to other source views from the target view by,
\begin{equation}
\Omega^{t\rightarrow s}_{p_i} = KM^{s}{M^{t}}^{-1}(\hat{\mathbf{D}}(p_i)\odot(K^{-1}\Omega^{t}_{p_i})),
\end{equation}
where $\hat{\mathbf{D}}(p_i)$ denotes the predicted depth of $p_i$, which is equal to $\hat{\mathbf{D}}(\mathbf{r})$. The ray $\mathbf{r}$ is emitted to the pixel $p_i$ from camera center.

%给定一个采样点p=(x,y)，它的支持域欧姆被定义为在一个带有大小为N的局部窗口，如:
\begin{equation}
\Omega_{p}=\left\{(x+x_p, y+y_p), x_p\in\left\{-N, 0, N\right\}, y_p\in\left\{-N, 0, N\right\}\right\}
\end{equation}

%像工作19,p2net， 我们定义我们的基于patch的多视角一致性photometric loss 为在支持域欧姆上的L1loss和structure similarity loss SSIM~\cite{SSIM}的组合，
With more robust depth warping and cross-view matching enabled by the support domain, like~\cite{monodepth2, p2net}, we propose to adopt a photometric loss over the support domain $\Omega_{p_i}$, which is the combination of an $\rm{L1}$ loss and a structure similarity loss $\rm{SSIM}$~\cite{ssim}. 

\begin{equation}
L_{SSIM}=SSIM(I_t[\Omega^t_{p_i}],I_s[\Omega^{t\rightarrow s}_{p_i}])
\end{equation}

\begin{equation}
L_{L1}=\left\|I_t[\Omega^t_{p_i}]-I_s[\Omega^{t\rightarrow s}_{p_i}]\right\|_1
\end{equation}

\begin{equation}
L_{ph}=\alpha L_{SSIM} + (1-\alpha) L_{L1},
\end{equation}
where \textit{support domain} $\Omega_{p_i}$ is defined as the local window surrounding the sampled point $p_i$. $I_t[\Omega_{p_i}^t]$ defines the pixel values at $\Omega_{p_i}^t$ in the target view $I_t$ via a bilinear interpolation. $\Omega_{p_i}^{t\rightarrow s}$ defines the region after warping the support domain $\Omega_{p_i}^{t}$ from the target view $I_t$ to the source view $I_s$. And $\alpha$ is a weighting factor that is set to $0.85$ empirically. By definition, $L_{ph}$ is patch-based and multi-view consistent.
%这里I_t[p_i]定义通过双线性插值的为p_i处的像素值在视角I_t，并且\alpha=0.85是一个加权因子。

%2.
Dense depth constraints are proved to be more beneficial to the geometry of neural radiance fields~\cite{dense_depth_priors}. Therefore, unlike P\({}^{\mbox{2}}\)Net~\cite{p2net}, we sample points directly from the whole image instead of the keypoints~\cite{dso}. Our experiments also show that dense sampling results in better performance than that of sampling from keypoints (See Sec. \ref{sub:ablation}). More importantly, in contrast to Dense Depth Priors~\cite{dense_depth_priors}, we achieve dense depth constraints free of any depth completion network which relies on external dataset training and have potential generalization problem.

% \textbf{Occlusion}. Following the pratices in~\cite{monodepth2}, we resolve the occlusion problem by using the minimum of the photometric error over the source images.
% %公式(featNet里的)
% \begin{equation}
% L^{\prime}_{ph}=\sum_{p_i}\min_{s\in V}L_{ph}(\Omega^{t}_{p_i}, \Omega^{t\rightarrow s}_{p_i}),
% \end{equation}
% where $V$ is a set composed of the previous and posterior source frames related to current target frames.
% 这里V是一个包含来源帧的集合,具体而言，是当前目标帧的前面的和后面的来源帧，

\subsection{Planar regularization with Superpixels}
\label{sec:plane-regularization}

% 室内场景包含了很多大面积的少纹理的区域，比如地板、墙和天花板等，这些区域使用基于patch的photometric values仍然不具有足够的判别性. Consequently the self-supervised depth learning of neural radiance field cannot be well guided by the patch-based photometric error.
Although patch-based photometric loss works well for textured regions, it fails to discriminate non-textured regions that are common in indoor scenes, such as floor, walls, tables and ceiling. We further observe that those non-textured regions are mostly planar. Therefore, how to inform \sysname{} of the planar constraints of a scene is the core concern. 
\begin{figure}[ht]
\centering
\includegraphics[width=1.0\linewidth]{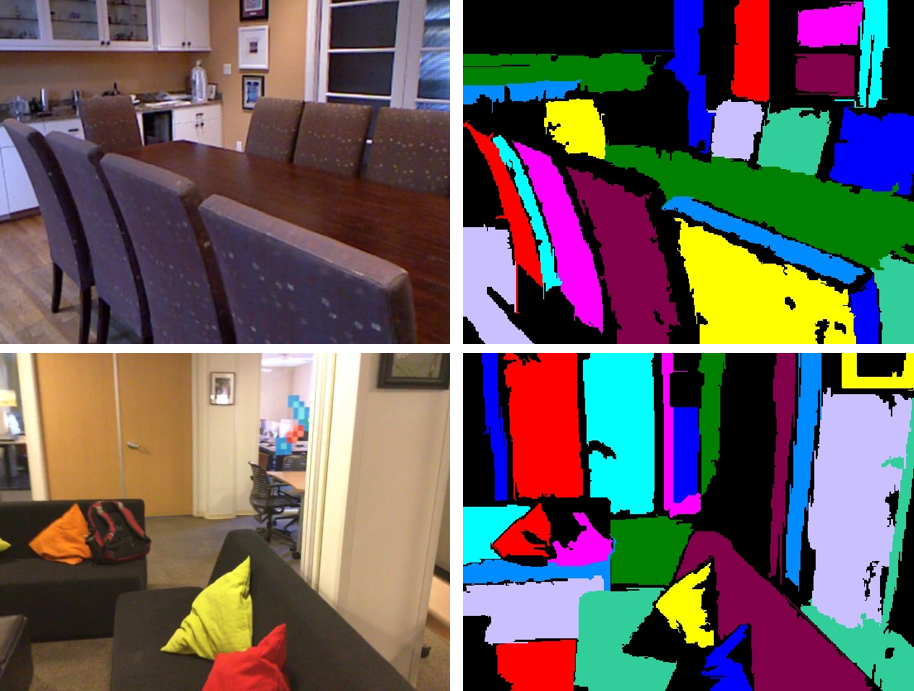}
\caption{Superpixel extraction (right) of two indoor images (left), colors represent different regions. We can see that most extracted regions are planes.}
\label{fig:superpixels}

\end{figure}
\begin{figure}[ht]
\centering
\includegraphics[width=1.0\linewidth]{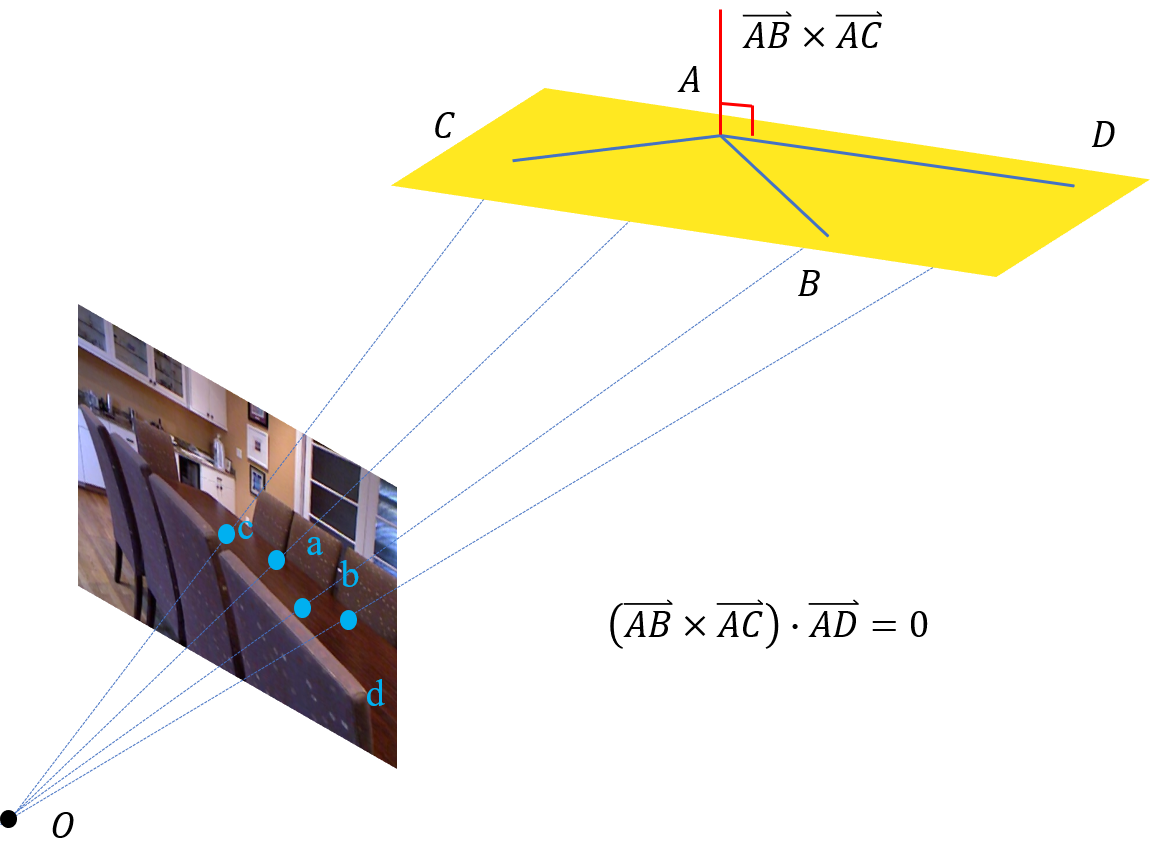}
\caption{The camestration process. We first re-project points $a, b, c, d$ in a 2D plane to the 3D coordinates, then enforce the cross product of $\mathop{AB}\limits^{\longrightarrow}$ and $\mathop{AC}\limits^{\longrightarrow}$ to be perpendicular to $\mathop{AD}\limits^{\longrightarrow}$. This figure is inspired by \cite{p2net}.}

% \wc{I think make the picture perspective (like the pictures in the patch-based multi-view consistent photometric loss in Fig.2) is better. Also, there are a lot of white regions at the top and right of this figure. Maybe you can make the planes of AB-AC and AB-AD colored.} 
% \wc{The colored image here may also be perspective}
% \cz{edited}
%\wc{cannot see a, b,c,d clearly due to color ambiguity}}
\label{fig:plane_reg}
\end{figure}

% \wc{I think this subsection is still a little simplified after I added the first paragraph, maybe you could add more details} 
Inspired by self-supervised depth estimation~\cite{plnet, p2net}, we aim to first identify 2D planes in input images by adopting the Felzenszwalb superpixel segmentation algorithm~\cite{graphseg}. Specifically, we extract superpixels from each view and define regions with area larger than a threshold as planes (We set it to be 1000 pixels in our experiments empirically) because those non-textured regions often span over a larger area. Fig.~\ref{fig:superpixels} provides examples that most of the segmented regions are planes.
%TODO:fig:superpixels
% \gycc{How is this value determined? You could say ``define regions with area larger than a threshold", and in your experiments you find 1000 works fine.}.\wc{edited}

Without specific regularization, NeRFs may fail to preserve the planar properties across different views, i.e. the depth map of planar regions is not flat. We propose to further impose the planar constraint to \sysname{} for non-textured regions using the planar consistency loss~\cite{plnet}. From each plane, we randomly sample 4 pixels, i.e., $a$, $b$, $c$ and $d$. With the rendered depth of \sysname{}, we then transform them to 3D points $A$, $B$, $C$, and $D$ in the camera coordinate with the following equation:

\begin{equation}
    P=\hat{\mathbf{D}}(p_i)\odot(K^{-1}p_i), p_i\in\left\{a, b, c, d\right\}, P\in\left\{A, B, C, D\right\},
\end{equation}
where $K$ denotes the matrix of camera intrinsic parameters. $p_i$ is the selected pixel and $P$ is the corresponding 3D point.

%如图所示，向量AB和向量AC之间的叉积和三角形ABC所在的平面垂直. D 应该和三角形ABC共平面，所以AD和ABxAC之间的叉积应该为0. , 

As shown in Fig.~\ref{fig:plane_reg}, the cross product of $\mathop{AB}\limits^{\longrightarrow}$ and $\mathop{AC}\limits^{\longrightarrow}$ should be perpendicular to the plane where $A$, $B$, $C$ and $D$ is located. Therefore, $\mathop{AD}\limits^{\longrightarrow}$ should be perpendicular to $\mathop{AB}\limits^{\longrightarrow}\times\mathop{AC}\limits^{\longrightarrow}$. The planar consistency loss is computed by,
\begin{equation}
L_{pc}=\frac{1}{N_p}\sum_{i=1}^{N_p}\left|\mathop{A_iB_i}\limits^{\longrightarrow} \times \mathop{A_iC_i}\limits^{\longrightarrow} \cdot \mathop{A_iD_i}\limits^{\longrightarrow} \right|,
\end{equation}
where $N_p$ denotes the number of 4-point sets we randomly select from planes.  

As shown in the experiments, \sysname{} achieves better performances both in terms of depth estimation and view synthesis for planar regions with the proposed plane regularization.

\subsection{Training}
\label{sec:training}

As introduced in DS-NeRF~\cite{dsnerf}, we also leverage the depth of sparse keypoints extracted by COLMAP~\cite{colmap1, colmap2} to supervise the geometry of the neural radiance field, which is view-consistent in nature.
\begin{equation}
    L_{sparse} = \sum_{x_i\in \mathbf{X}_j} w_i\left|\hat{\mathbf{D}}(\mathbf{r}_{ij}) - (\mathbf{M}_jx_i)\cdot[0,0,1]^{T} \right|^2,
\end{equation}
where the keypoint in camera $j$ is reprojected using camera extrinsic parameters $\mathbf{M}_j$ and then is projected onto its unit camera axis $[0,0,1]$. We also introduce a hyperparameter $w_i$ to adaptively adjust the weights of keypoints to reduce the negative influence of unreliable keypoints, determined by the reprojected error from COLMAP estimation. Supposing a 3D keypoint $x_i$ is visible in camera $j$, the reprojected error $e_{ij}$ is the distance in pixels between the camera coordinates $KM_jx_i$ and detected 2D keypoint in camera $j$. Thus, the confidence weight of a keypoint can then be measured by the total reprojected error $e_i=\sum_{j}e_{ij}$,
% 为了减少不可靠的关键点的影响，每个关键点的重投影误差被用来给关键点加权。假设给定了一个3D关键点x_i，在相机j中可见，重投影误差e_ij是投影出的相机坐标与在2D中检测出的关键点之间的像素距离；x_i的总共的重投影误差e_i=被用来度量这个点的置信权重：
% 公式
\begin{equation}
    w_i=exp(-(\frac{e_i}{\bar{e}})^2),
\end{equation}
where $\bar{e}$ denotes the mean absolute error of all the keypoints in a scene.
%这里 \overline
% $\bar{e}$ 表示一个场景中所有关键点的平均绝对误差.

We optimize the depth of neural radiance field by the weighted combination of patch-based multi-view consistency photometric loss, planar consistency loss and sparse depth loss from COLMAP as follows:
\begin{equation}
L_{\rm{Depth}}=\lambda_{ph}L_{ph}+\lambda_{pc}L_{pc}+\lambda_{sparse}L_{sparse}
\end{equation}
The overall loss function of \sysname{} is therefore:
\begin{equation}
L_{\rm{Total}} = L_{\rm{Depth}} + L_{\rm{Color}},
\end{equation}

% When we use
%在实际训练过程中，我们发现即使对L_sparse使用了上述加权的方法，这些噪声点云并没有消失，如后面的Ablations study的w/o warmup的深度图结果~\ref{fig:wo_warmup}展示的那样,还是会有很多noisy的floaters。仅仅调小权重不能完全解决，所以我们进一步尝试了一种更有效的方式。

\textbf{Warm-up Training}. During our training, we found that when we use fixed weights for $L_{sparse}$ over all iterations, the results were disturbed by the inaccurate points (See the noisy floaters in the depth map of the second column in Fig.~\ref{fig:wo_warmup}). Naively reducing $L_{sparse}$ would only diminish the benefits of the accurate points, so we adopt a warm-up training strategy. Specifically, we introduce the sparse depth loss item $L_{sparse}$ only in the first half of the training ($\lambda_{sparse}=0.05$). In the remaining iterations, we set $\lambda_{sparse}=0$ and let $L_{ph}$ and $L_{pc}$ refine the depths of pixels where noisy points are located. By setting $\lambda_{sparse}=0$, the noisy point clouds will not have an impact on NeRF anymore in the later training process. With warm-up training, we strike a balance of utilizing the sparse depth priors and avoiding noises of point clouds. 

% \gycc{This paragraph only describes the warm-up training strategy for $L_{sparse}$, therefore should be combined with your description about $L_{sparse}$. Also, there should be more analysis about why you use this warm-up strategy, ``To further reduce the negative influence" is not enough.}
%\cz{done}

% Total loss

\section{Experiments}
% \subsection{Experimental Setup}
% 参考Dense Depth Priors的补充材料网络结构解释

%我们使用Adam优化器，学习率设置为0.0005，我们以batch大小为1024的方式处理光线。为了公平，所有的消融实验和对比实验使用相同大小的MLP网络。

%辐射场总共优化了100k iterations.

%更多的细节可以在补充材料获得。

% 补充材料：，具体而言，我们的网络是使用NeRF原文~\cite{NeRF}中的结构，参考补充材料

% All $\lambda$ values and more implementation details are available in the appendix.

\subsection{Evaluation Metrics}
We use peak signal-to-noise (PSNR), the structural similarity index measure (SSIM)~\cite{ssim} and the learned perceptual image patch similarity (LPIPS)~\cite{lpips} to measure the quality of synthesized RGB novel views by comparing them with the ground truth. Besides, we also include two other metrics to demonstrate the effectiveness of our reconstructed geometry over previous methods. We use depth root-mean-square error (Depth RMSE) for measuring the predicted depth maps and the ground truth. Also, the Plane Mean Deviation is used to evaluate the flatness of planes for the predicted depth~\cite{plnet}. It is defined the distance of the measured point cloud to the fitted plane. We use the mean deviation to measure the flatness of planes for the predicted depth. Since the depth from NeRF is in a relative scale, different from the absolute ground truth depth from sensors. We therefore align the predicted depth to the ground truth according to conventional practice~\cite{median_scaling} because the deviation is scale-variant.

% \gycc{Why is the alignment necessary? Depth output by NeRF should be in correct scale.}.
% \cz{NeRF and COLMAP produce relative depths , but the ground truth depth is absolute. The scale is different, so we need scaling}

% \wc{better list the equation for plane mean deviation and plane flatness because they are as common as PSNR}
%这个deviation是被测量点云到拟合出的平面的距离。我们使用所有点的平均deviation(Mean Dev)来度量这个tolerance. 因为deviation是对尺度变化的，我们首先将预测出的深度和ground truth深度进行了对齐，依据传统的实践~\cite{median_scaling}

\subsection{Datasets}

To evaluate our method, We train and test our model on three multi-view indoor scene datasets in terms of novel view synthesis: ScanNet~\cite{scannet}, NYUv2~\cite{nyuv2} and SUN3D~\cite{sun3d}. We use only the RGB data for training and the ground truth depth are only used for evaluation.

% \gycc{Since these datasets all contain RGB-D data, you should be clear about (1) you only use RGB data to train your model; (2) you use the given ground truth depth (sensor output / render output) for evaluation.}
%\cz{edited}
%对于上述数据集中每个场景的训练视角我们跑了COLMAP来获取poses和稀疏点云，其中第8、16和24帧作为测试视角，剩下的作为训练视角(即训练视角不超过25帧).

For each scene of the datasets above, we run COLMAP~\cite{colmap1} over no more than 28 frames to obtain the poses and point clouds. We use eight sample scenes from each dataset respectively. Among the sampled frames, 8-th, 16-th and 24-th frames are used as the test views, and the rest are used as the training views.

% no more than \gycc{no less than?}
% \cz we use three test images.

% The SfM output is used only during training and is not used during inference.

%分辨率:
% scannet: 624x 468
% NYU:545x415
% SUN3D:624x 468
%我们从视频中每20帧中取出一帧
For ScanNet~\cite{scannet} and SUN3D~\cite{sun3d}, the image resolution is $468\times624$ after we down-sample and crop the dark borders from calibration. For NYUv2~\cite{nyuv2}, the image resolution is $545\times415$. %\wc{Sun3d resolution?}

\begin{table*}[]
    \vspace{20pt}
    \centering
    \begin{tabular}{lcccccccc}
        \hline
    {Dataset}&{Method} & Additional Training?& PSNR$\uparrow$ &SSIM$\uparrow$&LPIPS$\downarrow$&Depth RMSE$\downarrow$ & Plane Mean Dev$\downarrow$ \\ \hline
\multirow{4}{*}{NYUv2} & NeRF~\cite{nerf} & No &24.88 & 0.7899 & 0.2304 & 1.1512 & 0.0390    \\
&DSNeRF~\cite{dsnerf} & No & \underline{27.33} & \underline{0.8320} & 0.1943 & \underline{0.3604}  & 0.0347 \\
&Dense Depth Priors~\cite{dense_depth_priors} & Yes & 26.34 & 0.8219 & \underline{0.1774} & 0.5640 & $\bm{0.0250}$\\
&Ours  & No & $\bm{28.10}$ & $\bm{0.8561}$ & $\bm{0.1663}$ & $\bm{0.3113}$ & \underline{0.0301}\\
        \hline    
    
\multirow{4}{*}{SUN3D}&NeRF~\cite{nerf} & No & 17.99 & 0.6114 & 0.4451 & 1.6153 & 0.0375   \\
&DSNeRF~\cite{dsnerf} & No & \underline{22.26} & 0.7027 & 0.3430 & \underline{0.5761}  & 0.0386  \\
&Dense Depth Priors~\cite{dense_depth_priors} & Yes & 20.91 & \underline{0.7119} & \underline{0.3317} & 0.6866 & $\bm{0.0271}$ \\
&Ours  & No & $\bm{22.89}$ & $\bm{0.7627}$ & $\bm{0.2748}$ & $\bm{0.5287}$ & \underline{0.0327} \\
        \hline 

\multirow{4}{*}{ScanNet} & NeRF~\cite{nerf} & No &23.54 & 0.8017 & 0.2779 & 0.5527 & 0.0372   \\
&DSNeRF~\cite{dsnerf} & No & 24.25 & \underline{0.8190} & 0.2657 & 0.2500  & 0.0326  \\
&Dense Depth Priors~\cite{dense_depth_priors} & Yes & $\bm{24.84}$ & 0.8137 & \underline{0.2578} & $\bm{0.1929}$ & $\bm{0.0155}$ \\
&Ours  & No & \underline{24.67} & $\bm{0.8308}$ & $\bm{0.2480}$ & \underline{0.2298} & \underline{0.0270} \\
        \hline 
    \end{tabular}
    \caption{Comparisons with other methods. The performance of the thickened is the best. The underlined ranks the second in performance.}%Note: 我觉得第一列应该是指示帧编号的
\label{tab_cmp}
\end{table*}

\begin{table*}[]
    \vspace{20pt}
    \centering
    \begin{tabular}{lccccccc}
        \hline
    {Method} &Additioanl training?& PSNR$\uparrow$ &SSIM$\uparrow$&LPIPS$\downarrow$&Depth RMSE$\downarrow$ & Plane Mean Dev$\downarrow$ \\ \hline
w/o dense-sampling  &No& 27.18 & 0.8324 & 0.1822 & 0.3246 & 0.0329 \\
w/o patch &No& 28.03 & 0.8545 & 0.1671 & 0.3255  & 0.0312 \\
w/o warm-up training&No & 27.60 & 0.8456 & 0.1765 & 0.3382 & 0.0329\\
w/o sparse depth priors&No  & 27.42 & 0.8505 & 0.1734 & 0.4052 & 0.0350\\
w/o patch-match&No  & 27.56 & 0.8479 & 0.1772 & 0.3236 & $\bm{0.0300}$ \\
w/o plane regularization&No & 27.92 & 0.8498 & 0.1705 & 0.3308 & 0.0321  \\
full&No & $\bm{28.10}$ & $\bm{0.8561}$ & $\bm{0.1663}$ & $\bm{0.3113}$ & 0.0301 \\
        \hline       

    \end{tabular}

    \caption{Ablation studies on NYUv2 datasets. The performance of the thickened is the best.}
\label{tab_ablation_nyu}
\end{table*}

%Nerf:没有几何的约束
%DSNERF:只有Sparse点云的几何约束，缺少dense的几何约束
%Ours:有视角一致的dense几何约束，且在少纹理情况下有平面几何约束
%NerfingMVS:缺少视角一致性的效果

%Dense Depth Priors:需要大数据预训练、泛化性是一个问题、需要预设一个colmap detph和ground truth之间的scale、补全的深度缺少视角一致性。(补充材料里可以细说，实验depth loss weights=0.1)

%1.
%和Nerf, DSNerf,Ours(w/o latent), dense depth_prior(w/o latent)比较

%2. with latent:和dense depth prior比较

%3. 和dense depth priors结合的效果

%3. 我们的方法不需要大数据预训练，
%4.

\begin{figure}[h]
\centering
\includegraphics[width=1.0\linewidth]{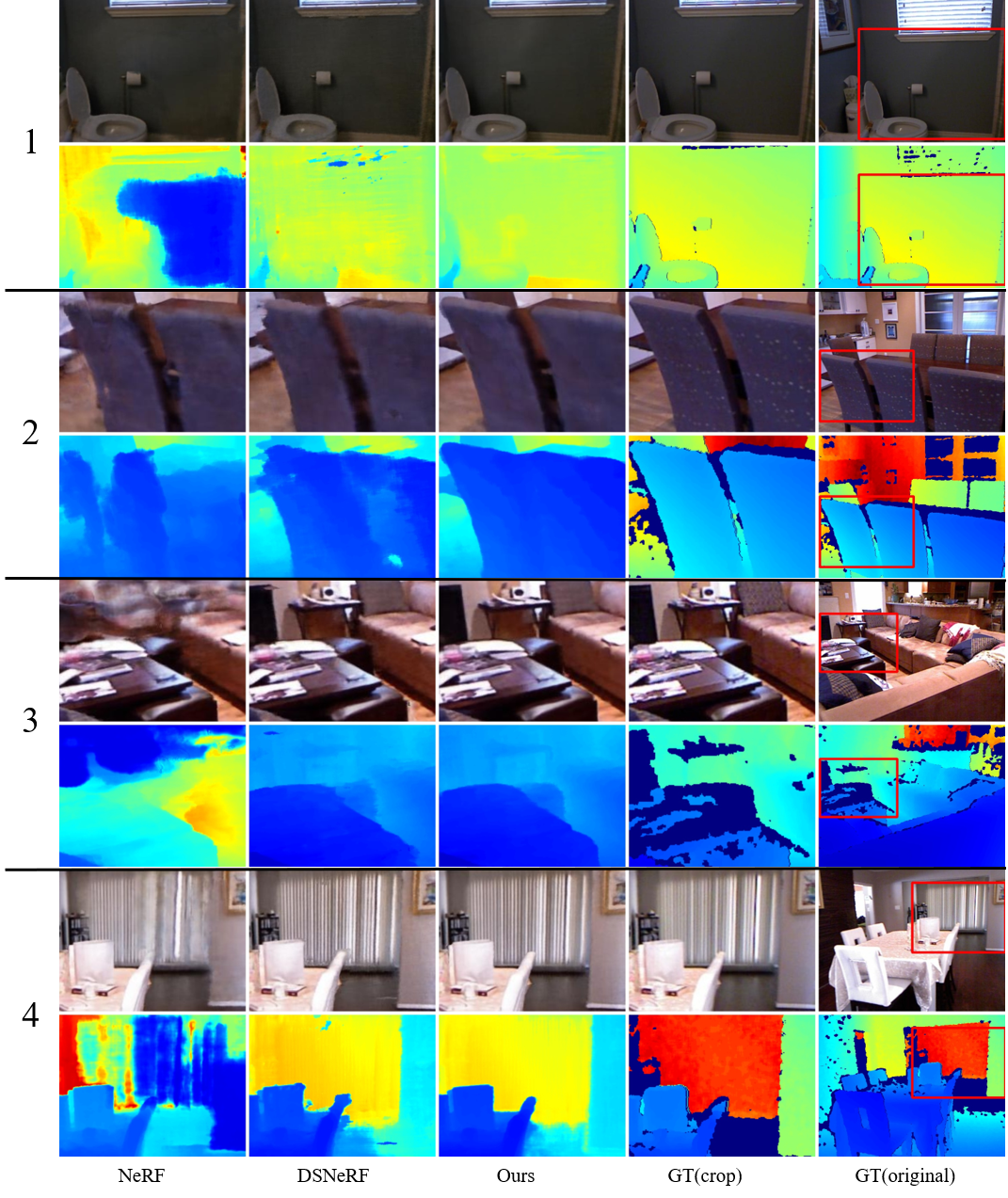}

% Rendered RGB and depth error for test views from three Matterport3D rooms next to the ground truth RGB and depth.
\caption{Comparisons with per-scene optimization method on the test views from NYUv2 scenes. "GT" denotes Ground Truth. "GT (original)" and "GT (crop)" mean the original and cropped ground truth respectively.}
\label{fig:nyu_cmp_without}
\end{figure}
% \wc{The 1,2,3,4 are not clear enough}
% \cz{edited}

\begin{figure}[!h]
\centering
\includegraphics[width=1.0\linewidth]{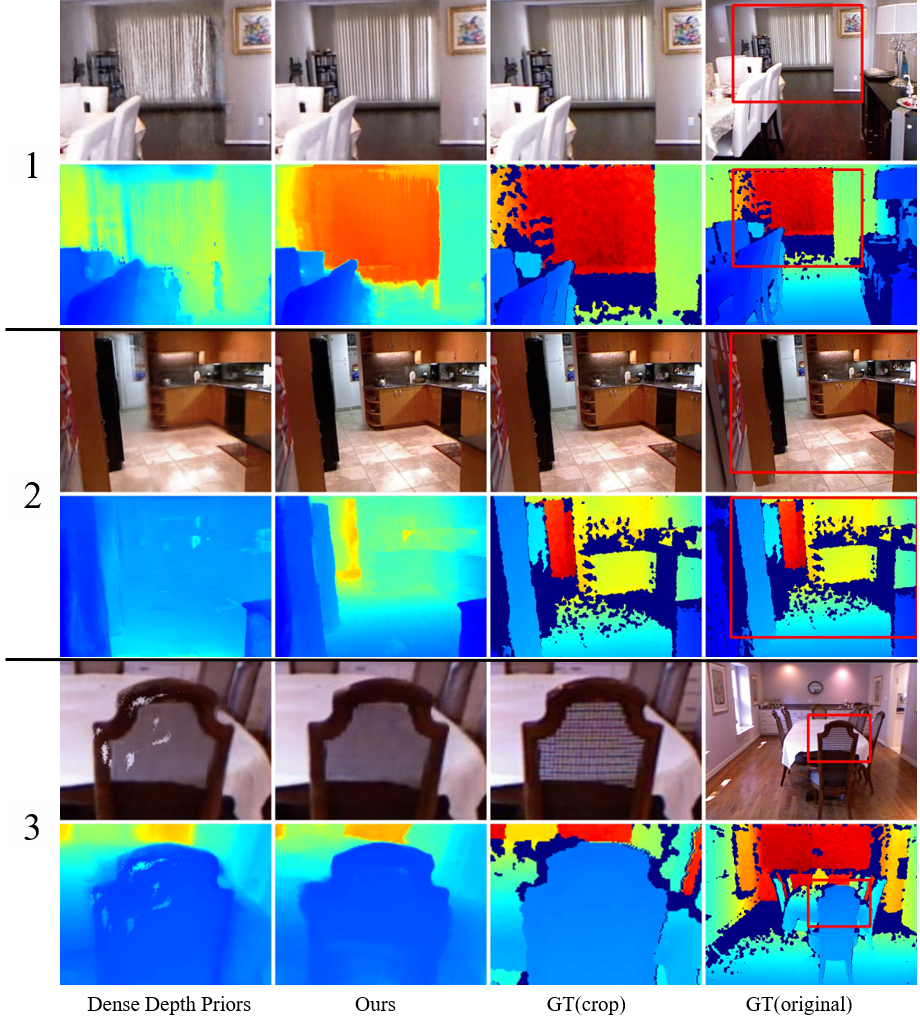}
\caption{Comparisons with data-driven method on the test views from NYUv2 scenes. "GT" denotes Ground Truth. "GT(original)" and "GT(crop)" mean the original and cropped ground truth respectively.}
\label{fig:nyu_cmp_with}
\end{figure}

\begin{figure}[!h]
\centering
\includegraphics[width=1.0\linewidth]{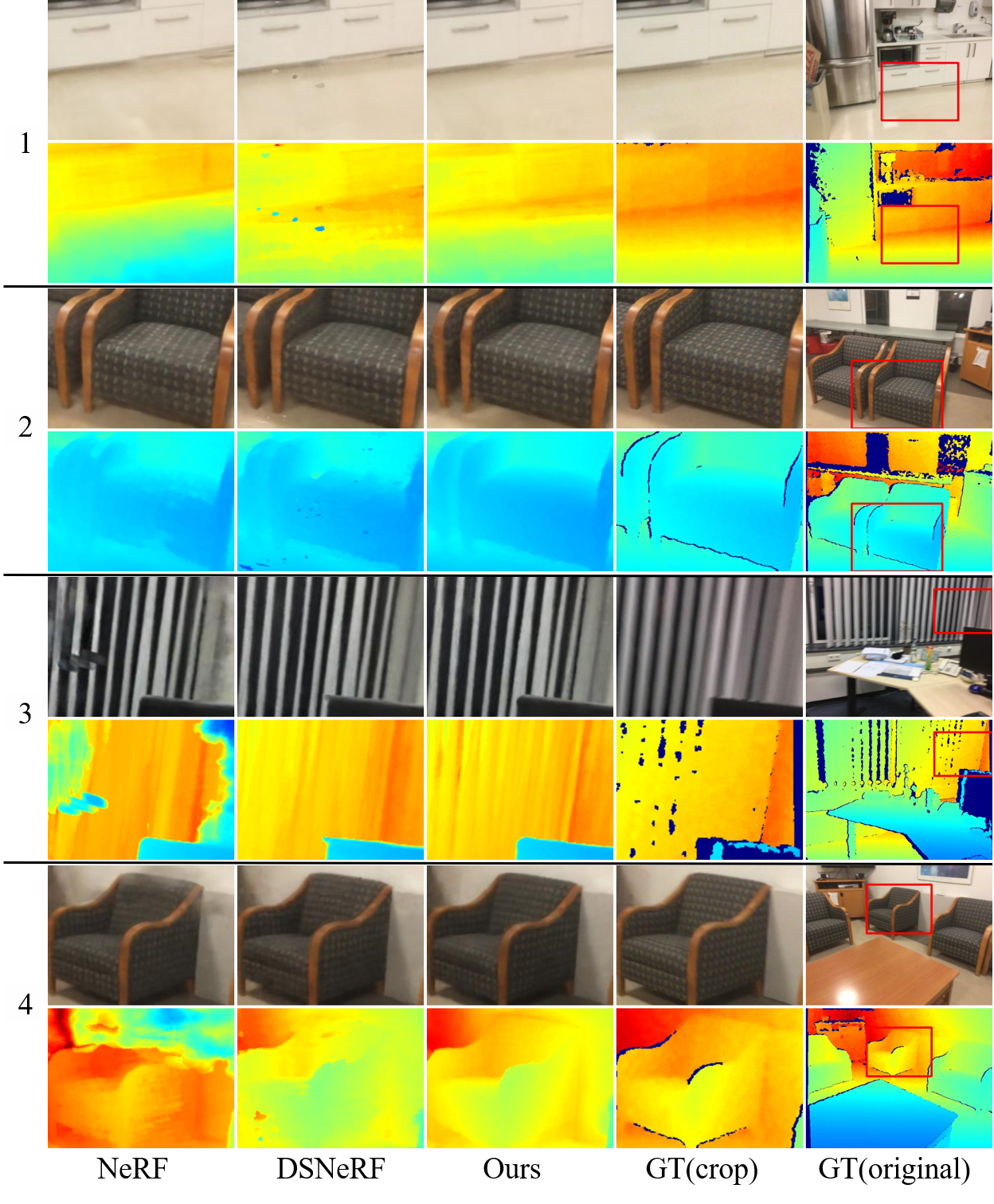}
\caption{Comparisons with per-scene optimization method for the test views from ScanNet scenes. "GT" denotes Ground Truth. "GT(original)" and "GT(crop)" mean the original and cropped ground truth respectively.}
\label{fig:scannet_cmp_without}
\end{figure}

\begin{figure}[h]
\centering
\includegraphics[width=1.0\linewidth]{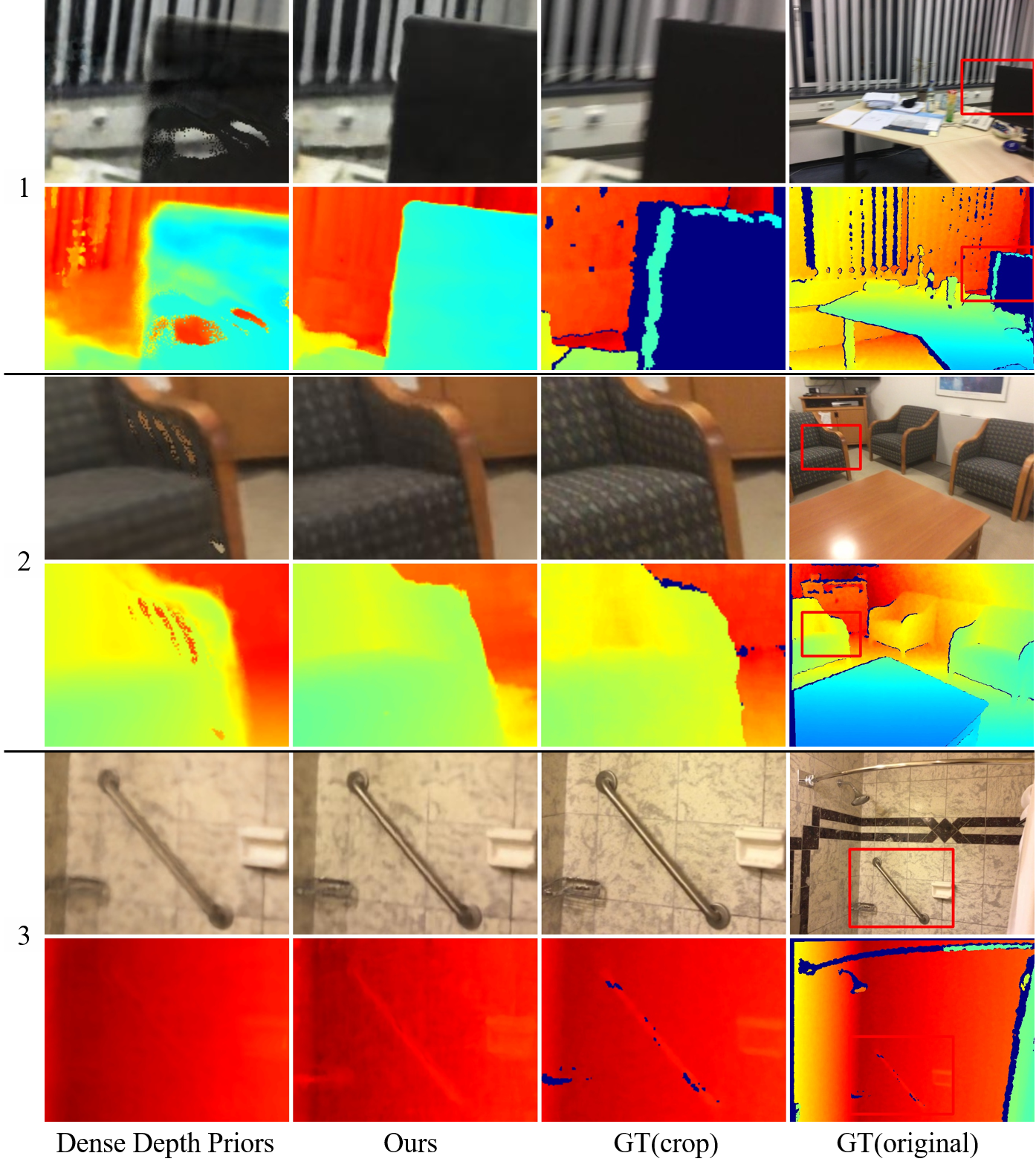}
\caption{Comparisons with data-driven method for the test views from ScanNet scenes. "GT" denotes Ground Truth. "GT(original)" and "GT(crop)" mean the original and cropped ground truth respectively.}
\label{fig:scannet_cmp_with}
\end{figure}

\subsection{Implementation Details}
% \gycc{Put this paragraph after ``Datasets".}
%\cz{done}
We set $\lambda_{ph}=0.025$ and $\lambda_{pc}=0.025$ and $\lambda_{sparse}=0.05$ in all the scenes of each dataset. $N$ is set to 2, and we take the previous two frames and the posterior two frames of the target view as the source views, which is the same as~\cite{p2net}. We use the Adam optimizer~\cite{adam} with learning rate 0.0005 and sample rays in batches of 1024. The radiance field is optimized for 100k iterations. We use the same MLP architecture in all experiments as NeRF~\cite{nerf} for fair comparison.

\subsection{Comparison with Existing Methods}
We compare \sysname{} with existing methods for novel view synthesis given sparse inputs in the indoor scenes. The baselines include two categories: the methods trained by per-scene optimization without external data and the methods with data-driven depth priors (we call data-driven methods).

% mainly fall into two categories: per-scene optimization without pretraining and pretraining with per-scene optimization.

% Compared to methods that utilizes external data to train their depth estimation networks (we call pretrain-based methods)

% \gycc{``pretraining" here is not very appropriate (also see my comments in the introduction). It could be confused with methods like PixelNeRF. Consider ``Methods with data-driven depth priors"?}

\subsubsection{Comparison with per-scene optimization Methods}

%我们将我们的方法和NeRF、最近的工作使用稀疏点云的工作的NeRF，也就是DS-NeRF~\cite{dsnerf}作了比较。NeRF,DSNeRF，我们的方法都在ScanNet、SUN3D、NYUv2上跑了。 大量的实验结果(Tab)表明，我们的方法在所有指标上都超过NeRF和DSNeRF。
Per-scene optimization methods include vanilla NeRF~\cite{nerf} and DSNeRF~\cite{dsnerf} that uses sparse point clouds from SfM for depth supervision.
We ran NeRF, DSNeRF, and \sysname{} on the three datasets, ScanNet, SUN3D and NYU. The quantitative results (as shown in Tab.~\ref{tab_cmp}) shows that our method outperforms NeRF and DSNeRF in all the metrics. 
The visualized results of comparisons on ScanNet and NYUv2 are listed in Fig.~\ref{fig:scannet_cmp_without} and Fig.~\ref{fig:nyu_cmp_without}. The visualized results of comparisons on SUN3D can be seen in the appendix. NeRF produces the worst results because its geometry is extremely unconstrained (See Fig.~\ref{fig:scannet_cmp_without} and Fig.~\ref{fig:nyu_cmp_without}). DSNeRF has only sparse depth priors from COLMAP, it often produces artifacts in the depth-unconstrained areas. Wrong color and geometry are produced by DSNeRF, e.g., visible in the chairs (Example 2 in Fig~\ref{fig:nyu_cmp_without}). In contrast,
\sysname{} renders more accurate depths and colors because \sysname{} learns two structural hints which supervise the geometry of NeRF at textured and non-texture regions respectively. Besides,
We found that \sysname{} is more robust to the outliers in the sparse depth input in the edge of the window (Example 1 in Fig.~\ref{fig:nyu_cmp_without}), while the unnecessary floaters are very obvious for DSNeRF.

% \gycc{Need more analysis. Refer to the figure to see specific differences between the results of your method and baseline methods. Does your method outperform DS-NeRF in areas where there are COLMAP keypoints?}
% \cz{edited}

% \wc{explanation for DS-NeRF}
%DS-NeRF因为只有稀疏的深度先验，在深度没有受到约束的区域会产生artifacts。我们的方法对于少纹理区域和一般的带纹理的区域的几何都做了约束，在几何和合成上效果都比DS-NeRF好。

%列出NYU、SUN3D、ScanNet图片(至少一页对比图片)，然后一一列举说明我们的方法相对于NeRF和DSNeRF的优点。

% 我们的方法相对于NeRF和DSNeRF在几何上有较明显的改善
% 
%列出patchmatch改善的区域(边缘、带有纹理的区域)

%列出plane改善的区域(少纹理区域)

%列出warmup改善的区域(floaters, noises)

\subsubsection{Comparison with Data-driven Methods}

We also compare our method to the recent work (Dense Depth Priors~\cite{dense_depth_priors}) that uses the depth completion network to complete depths from sparse depth inputs and then uses them to supervise the geometry of NeRF. Different from our method, the depth completion network of Dense Depth Priors needs to be pretrained on a large-scale indoor dataset ScanNet in a supervised way. The completed dense depths for new scenes are pre-calculated by the depth completion network and used to supervise the depth of NeRF when training NeRF on the new scenes. Like NeRF~\cite{nerf}, our method is only optimized on the specific scene without additional training. Note we didn't include a comparison with another relevant work NerfingMVS\cite{nerfingmvs} because it fails on most scenes for reasons we explained in the appendix.

The results (Tab.~\ref{tab_cmp}) show that, with the dataset prior, Dense Depth Priors is better than \sysname{} on ScanNet in terms of Depth RMSE. But \sysname{} still has comparable PSNR and even better SSIM and LPIPS compared with Dense Depth Priors. We also find that our method is more view-consistent than Dense Depth Priors (shown in Example 1 and 2 of Fig.~\ref{fig:scannet_cmp_with} because the predicted depths of the depth completion network are not view-consistent. Besides, we found that the depth estimations from data-driven models for Dense Depth Priors shows less detailed depth and color (such as the metal handrail in Example 3 of Fig.~\ref{fig:nyu_cmp_with}).

% \gycc{Refer to specific areas where Depth Depth Priors / NerfingMVS produces inaccurate depth estimations from pretrained models.}
%\cz{done}

Moreover, the performance of Dense Depth Priors degrades significantly on SUN3D and NYUv2 as shown in Tab.~\ref{tab_cmp}, mainly due to the generalization issue of the depth completion network. Fig.~\ref{fig:nyu_cmp_with} shows that our method doesn't suffer from the generalization problem and surpasses Dense Depth Priors in every metric other than the Plane Mean Dev. The visualized results of comparisons on SUN3D can be found in the appendix.

% \begin{table}[]
%     \vspace{20pt}
%     \centering
%     \begin{tabular}{ccccc}
%         \hline
%     {Method} & PSNR$\uparrow$ &SSIM$\uparrow$&LPIPS$\downarrow$&Depth RMSE$\downarrow$  \\ \hline
% DenseDepthPriors & - & - & - & - \\
% Ours & - & - & - & -\\
%         \hline       
%     \end{tabular}
%     \caption{Comparisons with DenseDepthPriors on NYUv2 and ScanNet. "-" denotes that the method fails in some cases.}%Note: 我觉得第一列应该是指示帧编号的
% \label{tab_cmp_dense_nyu}
% \end{table}
% 放一张比较的图像(和DSNeRF、NeRF的比较图像,NYU和ScanNet),这些方法不需要数据集先验

% 放一些比较NerfingMVS的例子和表格（可以放补充材料里）

% 再放一张和DenseDepthPriors的比较图像(NYU和ScanNet)

\subsubsection{Implementation Details of Baselines}

\noindent\textbf{DSNeRF} 
The depth loss weight was set to 0.1 in all the scenes of each dataset as recommended in ~\cite{dsnerf}.

\vspace{2pt}

\noindent\textbf{Dense Depth Priors} The depth loss weight was set to 0.008 in all the scenes of each dataset as recommended in~\cite{dense_depth_priors}. The depth completion network is pretrained on ScanNet~\cite{scannet}. We use the official released weights from Github\footnote{https://github.com/barbararoessle/dense\_depth\_priors\_nerf} to predict the dense depth maps in all of our experiments. The radiance field of Dense Depth Priors is trained on each scene with the supervision of completed depth maps.

\subsection{Ablation Study}
\label{sub:ablation}

% \gycc{This subsection is the \textbf{CORE} of your experiments (not the comparisons). Currently it's too short and lacks detailed analysis. Try to explain \textbf{case-by-case} how your proposed constraints (especially patch-match and plane regularization) work, when they work best, and when they could potentially fail. Spend more pages on the analysis of your unique contributions since they are what the readers/reviewers really care about (now it seems all evaluations are of the same importance).}
% \cz{}
% without
% 我们在NYUv2上执行了消融实验,为了验证添加的部件的有效性。如表格所示，大量的结果说明我们模型完整的版本得到了最好的效果。具体的可视化效果如图所示。
We conduct ablation studies on NYUv2 to further validate the effectiveness of the different components in \sysname{} for view synthesis and depth estimation. The quantitative results can be found in Table~\ref{tab_ablation_nyu}.

\begin{figure}[!h]
\centering
\includegraphics[width=1.0\linewidth]{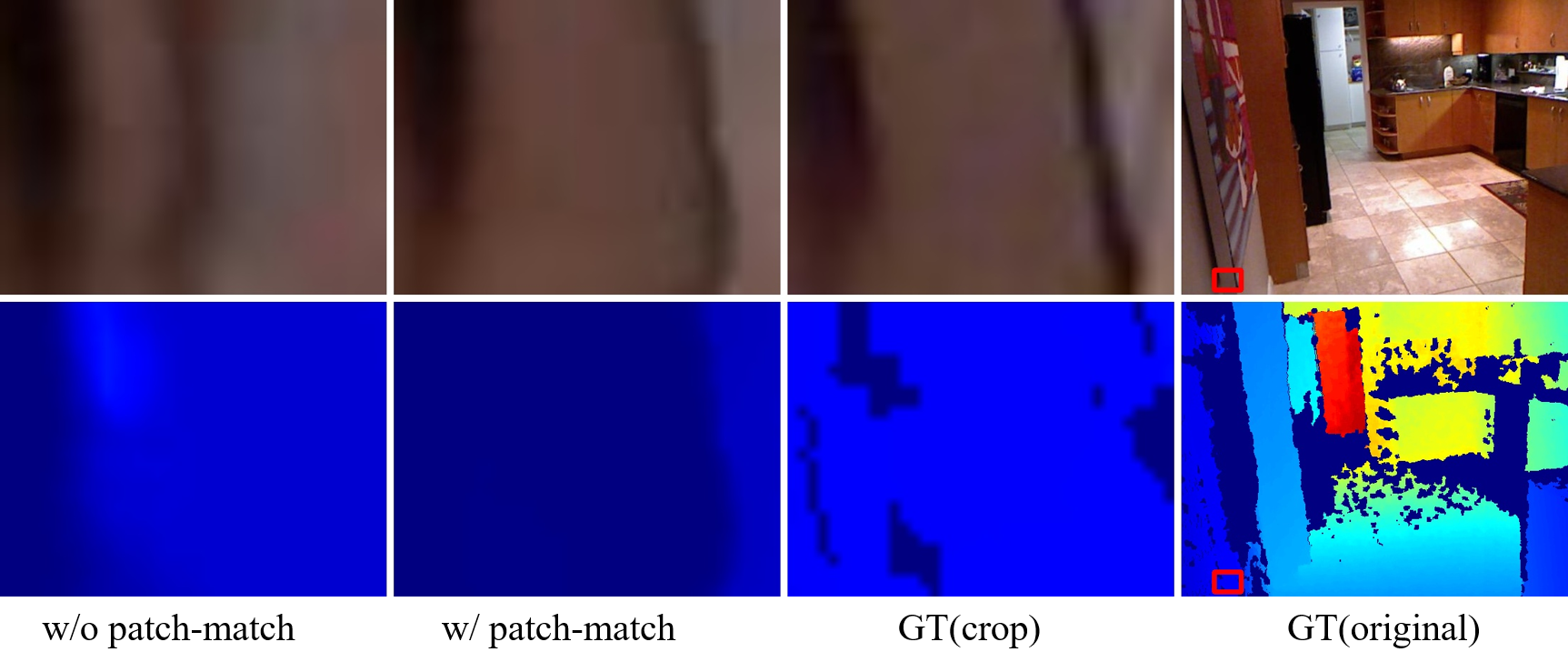}
\caption{w/o patchmatch and w/ patchmatch.}
\label{fig:wo_patchmatch}
\end{figure}

\begin{figure}[!h]
\centering
\includegraphics[width=1.0\linewidth]{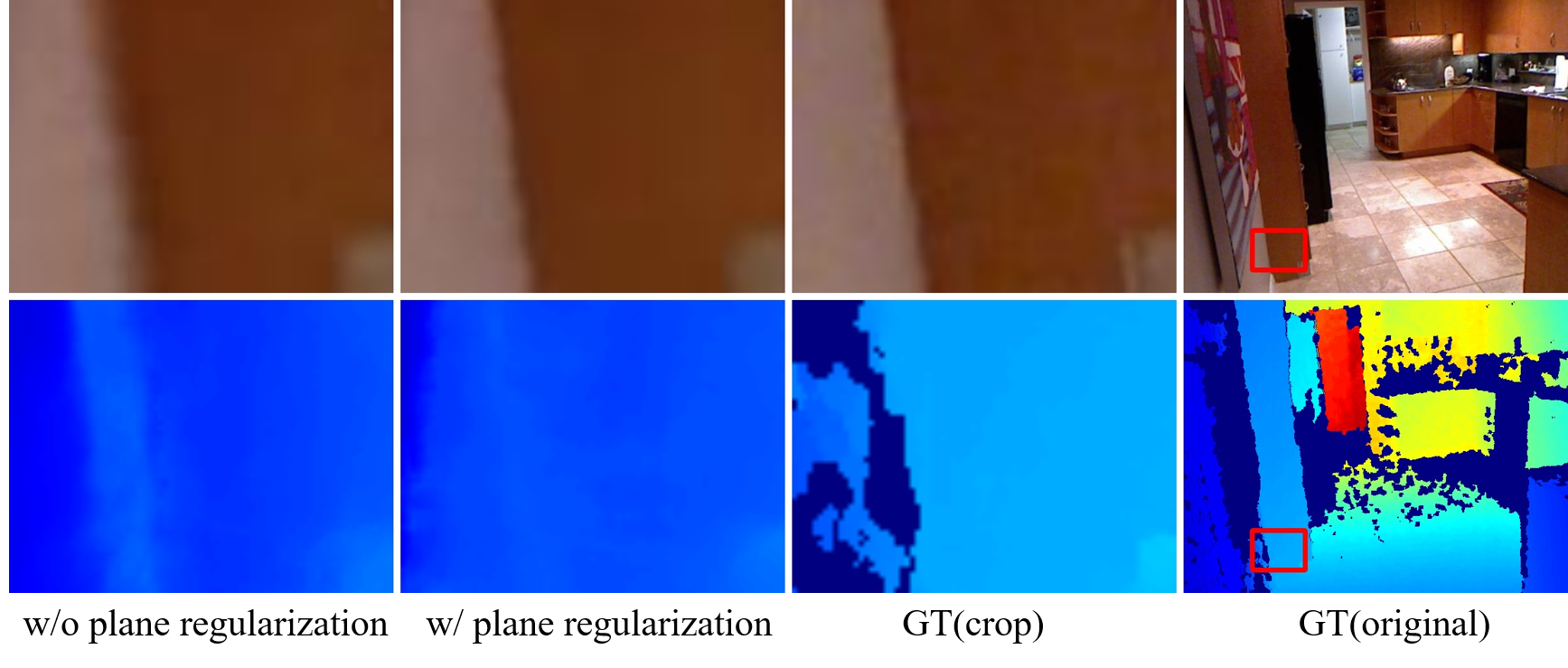}
\caption{w/o plane regularization and w/ plane regularization.}
\label{fig:wo_plane_reg}
\end{figure}

% 省略patch-match

\begin{figure}[!h]
\centering
\includegraphics[width=1.0\linewidth]{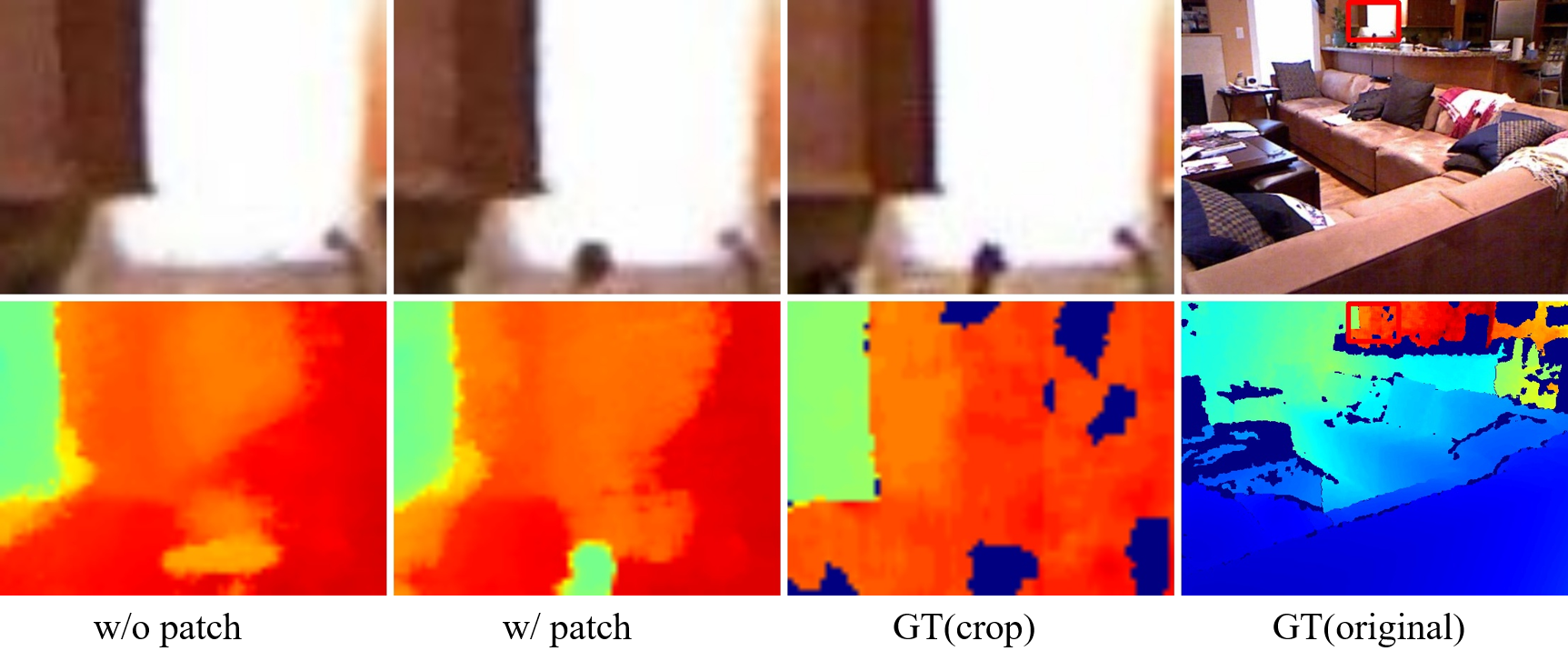}
\caption{w/o patch and w/ patch.}
\label{fig:wo_patch}
\end{figure}

\begin{figure}[!h]
\centering
\includegraphics[width=1.0\linewidth]{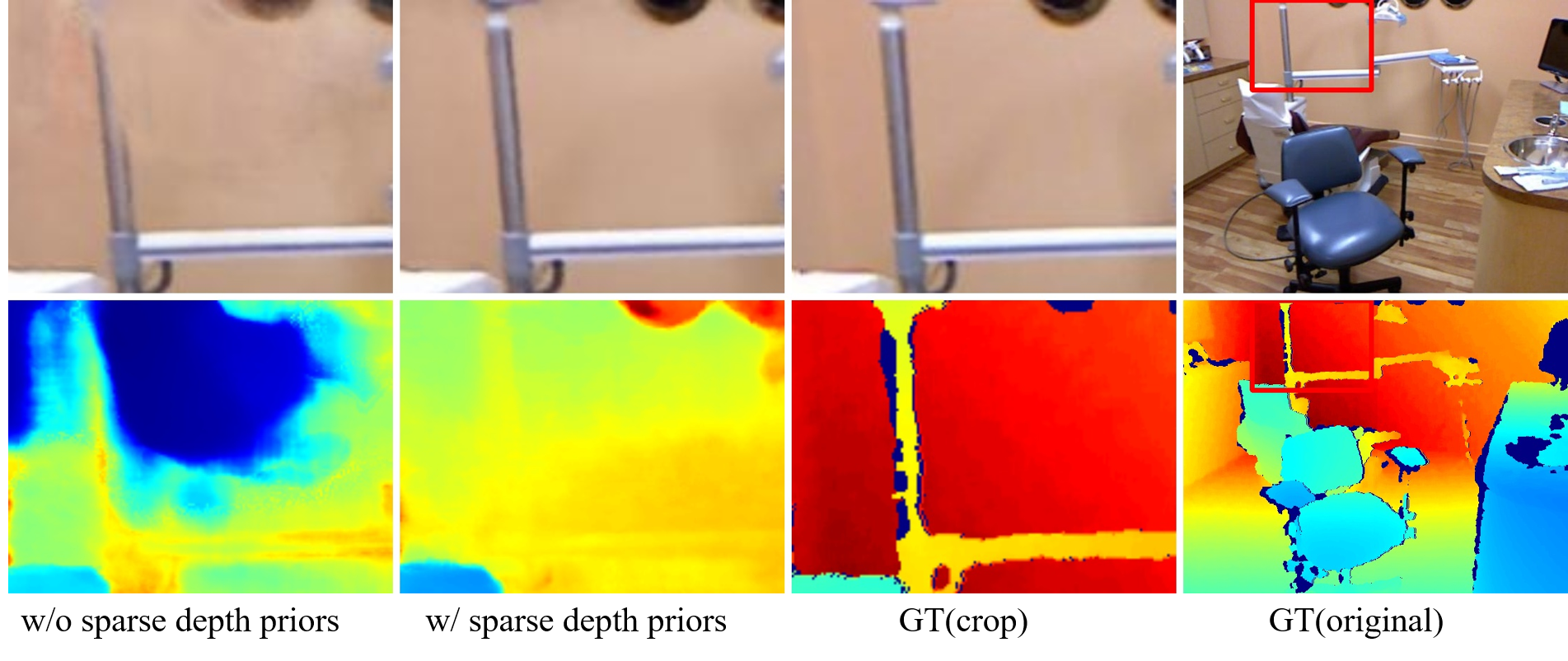}
\caption{w/o sparse and w/ sparse.}
\label{fig:wo_sparse}
\end{figure}

\begin{figure}[!h]
\centering
\includegraphics[width=1.0\linewidth]{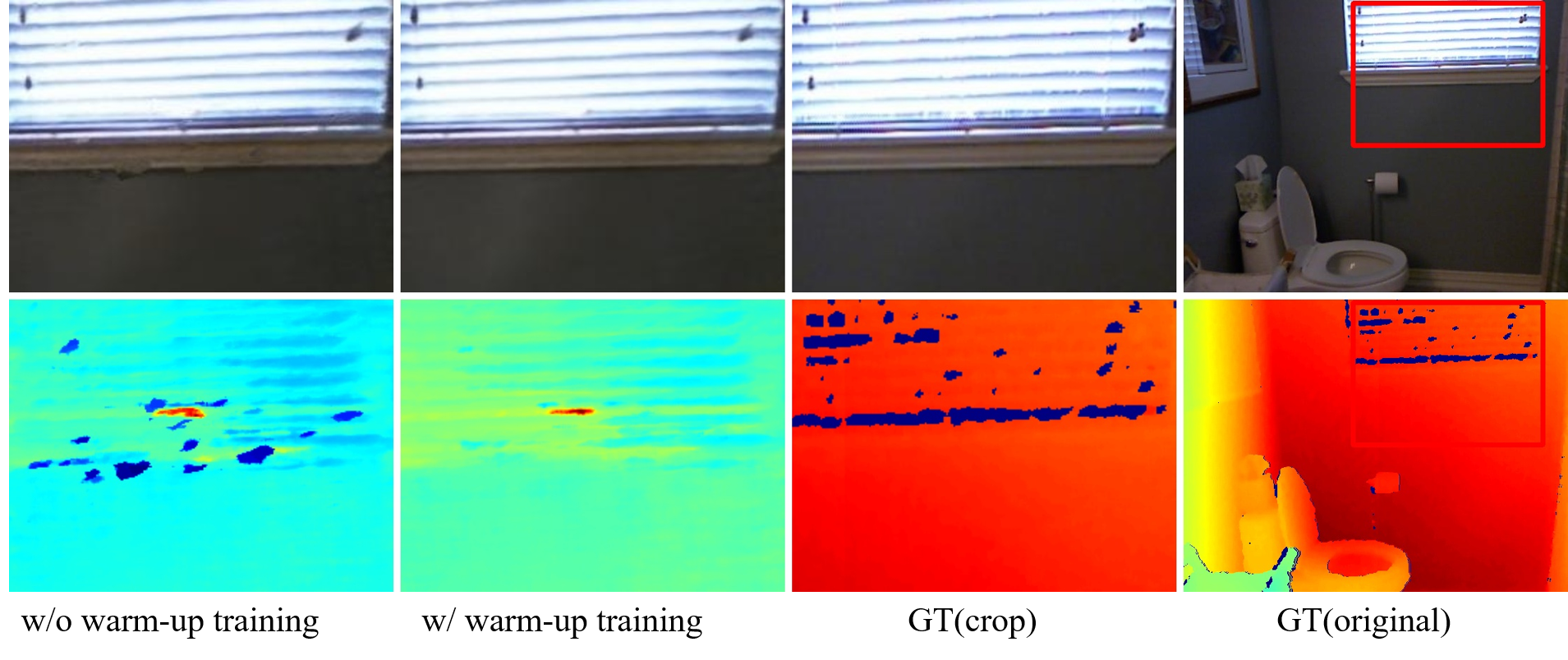}
\caption{w/o warm-up training and w/ warm-up training.}
\label{fig:wo_warmup}
\end{figure}

\begin{figure}[!h]
\centering
\includegraphics[width=1.0\linewidth]{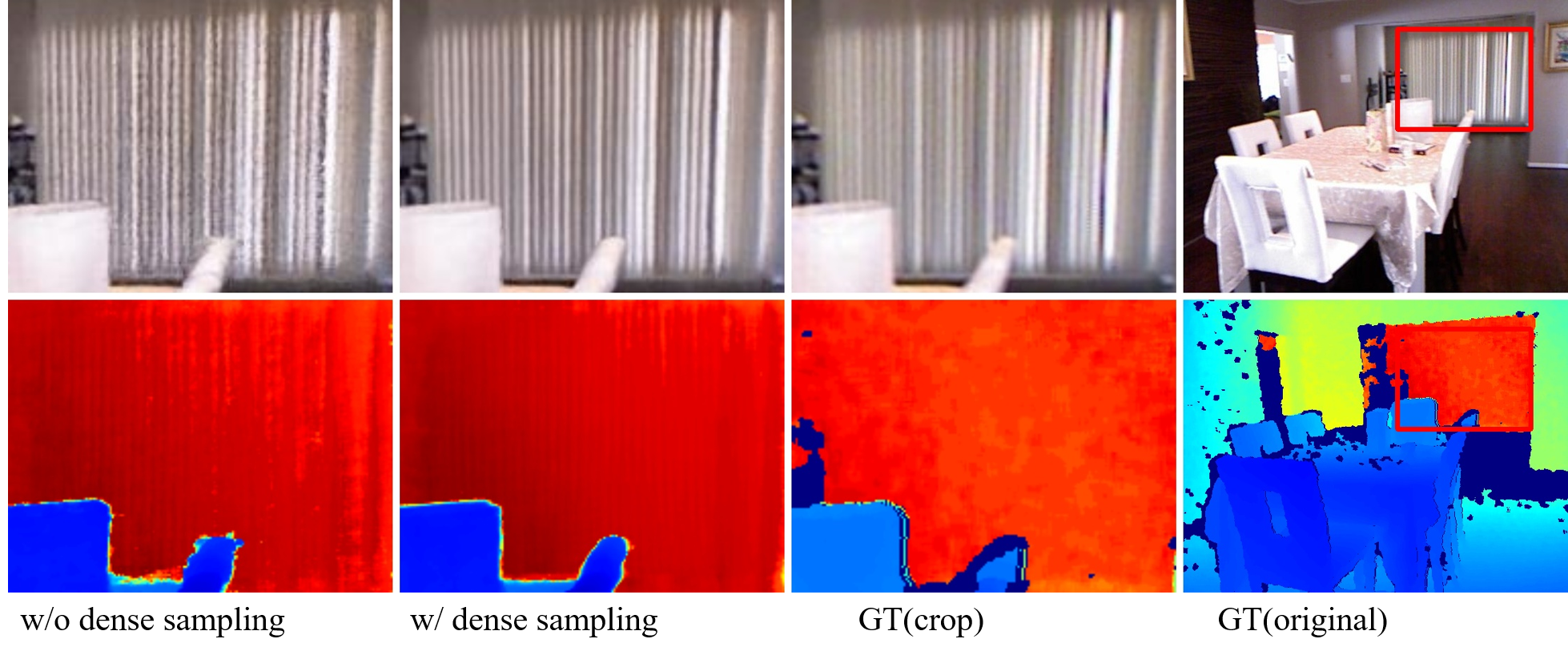}
\caption{w/o dense sampling and w/ dense sampling.}
\label{fig:wo_dense_sampling}
\end{figure}

\vspace{3pt}
\noindent\textbf{Patch-match} Omitting patch-match leads to inaccurate depth and color in high-frequency areas. The ability of NeRF to model the appearance of view-dependent appearance leads to the ambiguity between its 3D shape and radiance~\cite{nerf++}. With multi-view consistency, patch-match improves the geometry of textured regions and reduces the artifacts in the edges such as the edges of the billboard shown in Fig.~\ref{fig:wo_patchmatch}. When it is omitted, the black edge appears in a wrong place and its shape is wrongly estimated because it lacks the corresponding geometry constraints. $L_{sparse}$ only provides the depth constraints ats sparse keypoints and $L_{pc}$ at non-textured regions. Patch-based multi-view consistency is necessary for the textured regions other than keypoints.

% The ability of NeRF to model the appearance of view-dependent appearance leads to the ambiguity between its 3D shape and radiance~\cite{nerf++}. To reduce the shape-radiance ambiguity,
% we leverage multi-view consistency to supervise the depth of every pixel for each view.

% with view-consistent constraints for the geometry 

%patch-match通过基于patch的多视角的一致性，约束了场景中带有纹理的边缘区域几何准确性，当我们消除这个约束时，我们会发现图像中细小的瓶子因为缺少了对应的几何约束，在场景中没有重建出来，采取这种多视点的一致性，我们发现对于NeRF的几何和合成，都会更加鲁棒。

\vspace{3pt}
\noindent\textbf{Plane Regularization} 
%去除plane regularization之后，少纹理的区域的几何会明显变差，对应的颜色也会变差一些。The Depth RMSE 和Plane Mean Dev and Flatness Ratio会明显变差。 PSNR也会下降一些。
Removing plane regularization causes the geometry of textureless regions becomes less constrained, leading to less sharp edges in RGB, as shown in Fig.~\ref{fig:wo_plane_reg}. As a result, the Depth RMSE and Plane Mean Dev degrades. In detail, after we remove the plane regularization, the left white area is mistaken for two planes and the colored edge between the white and brown area is less clear. Since the white area corresponds to a plane in \sysname{} represented by a superpixel, our proposed planar consistency loss enforces flat geometry in this region, which reduces the artifacts in both the color and depth. It makes up for the insufficiency of patch-match in the non-texture regions.

% \wc{what ambiguity? shape and radiance?}
% \cz{remove the ambiguity.}

\vspace{3pt}
\noindent\textbf{Patch-based vs. Point-based Photometric Loss}
We replace the patch-based multi-view consistency photometric loss with the point-based one. It can be easily seen from Fig.~\ref{fig:wo_patch} that patch-based loss leads to a robust rendering quality (the tiny bottle disappears in the w/o patch setting).

%The PSNR, SSIM, LPIPS, DepthRMSE等指标各下降了多少

\vspace{3pt}
\noindent\textbf{Sparse Depth Priors}
Excluding the sparse depth priors, we observe that NeRF is more likely to fall into the local optimal as shown in Fig.~\ref{fig:wo_sparse}. Therefore, although sparse depth priors from COLMAP contain many noises, our experiments show that they are still indispensable for NeRF with sparse views. 

\vspace{3pt}
\noindent\textbf{Warm-up Training} 
Omitting the warm-up training strategy and using the same $\lambda_{sparse}$ across all iterations cause the noises of point clouds to be much more obvious in RGB and depth maps (See the floater artifacts in Fig.~\ref{fig:wo_warmup}). The artifacts are perfectly removed with the warm-up strategy.

\vspace{3pt}
\noindent\textbf{Dense-sampling} In this experiment, we sample at the keypoints extracted by~\cite{dso} in patch-match to supervise the depth of NeRF. We observe that sparse keypoint sampling leads to under-constrained depth and worse rendering results (as shown in Fig.~\ref{fig:wo_dense_sampling}). Dense sampling instead helps patch-match supervise the geometry of most regions.

% without Occlusion
% without dense-sampling
% without patch. -based presentation

% \section{Limitations}
%我们的方法允许在少视角下对输入图像进行新视角合成。但是我们的方法并没有解决NeRF的长时间优化问题和很慢的渲染效率问题。，和其他方法~\cite{nerf, dsnerf, dense_depth_priors}一样，因为场景表面只被少数视角观察到我们的方法也无法很好处理view-dependent效应。
% Although our method allows novel view synthesis with sparse input views, the limitations of NeRF such as long optimization time and slow rendering still remain in our method. 
% Like other methods~\cite{nerf, dsnerf, dense_depth_priors}, since the surfaces are observed by only a few views, view-dependent effects are limited.

% Our method reconstructs flatter planes with NeRF than those methods without pretraining. But it is still inferior to Dense Depth Priors which is pretrained on a large-scale indoor dataset in a supervised way.

\section{Conclusion and Future Work}
% The conclusion goes here.
% 本文提出了一种带有自监督约束的神经辐射场，用于少视角输入的室内场景。我们的方法对于少纹理区域利用平面正则化来约束，对于有纹理区域利用patch-match来优化几何，对于带有noisy点云的区域，采取了warm-up的策略来改善。相对于之前的方法，我们的方法在多个室内场景数据集上具有更好的视角合成效果。
This paper proposes \sysname{}, neural radiance field with self-supervised depth constraints for indoor scene novel view synthesis with sparse input views. We are the first to apply structural hints from multi-view inputs to NeRF for view synthesis and geometry estimation, specifically, patch-match and plane regularization to constrain the depth of textured and textureless regions respectively. In this way, it learns a view-consistent geometry with dense depth constraints. Most importantly, we doesn't have the generalization problem which occurred in data-driven methods, \eg Dense Depth Priors~\cite{dense_depth_priors}. Besides, we adopt a warm-up training strategy to reduce the influence of noisy point clouds from Structure-from-Motion. \sysname{} outperforms state-of-the-arts without additional training~\cite{nerf,dsnerf} both in depth estimation and novel view synthesis. In terms of comparison to data-driven methods, \ie Dense Depth Priors~\cite{dense_depth_priors}, it still achieves comparable performance on the pretrained dataset (ScanNet) and superior performance on other datasets (SUN3D and NYUv2). \sysname{} raises the upper bound of rendering quality of NeRF without external data given sparse input views. Our work also motivates future research to further exploit the structural hints in multi-view inputs for view synthesis and other related tasks.

% \wc{Most importantly, we doesn't have the generalization problem which occurred in data-driven methods, \eg Dense Depth Priors~\cite{dense_depth_priors}}
%\cz{edited}

% \wc{check this: We are the first to apply structural hints from multi-view inputs to NeRF for view synthesis and geometry estimation}.
%\cz{edited}
% \wc{Our work also motivates future research to further exploit the structural hints in multi-view inputs for view synthesis and other related tasks.}
%\cz{edited}

% In conclusion, without external data, \sysname{} achieves the state-of-the-art performance in the indoor scenes for novel view synthesis with sparse input views. It does not need to consider the generalization problem which often occurs in data-driven methods, \eg, Dense Depth Priors~\cite{dense_depth_priors}. With two structural hints of indoor scenes, \wc{I have incorporated this in the last paragraph}

% \gycc{Need more general statements about how your method might benefit the community or industry.}
% \cz{add the words after "In conclusion,..." as shown above}

Limitations of \sysname{} include limited view-dependent effects since the surfaces are observed by only a few input views, which also happens in other baselines~\cite{nerf, dsnerf, dense_depth_priors}. Also, our method in plane reconstruction is still slightly inferior to supervised data-driven methods, albeit we already surpassed per-scene optimization ones significantly. In the future, We will consider how to model the view-dependent effect in the sparse input setting. We would also investigate how to incorporate the rich priors of indoor datasets, possibly with a more generalized NeRF trained across large-scale datasets.
%Future work
% 我们未来考虑如何保证少视角输入下的view-dependent的效果。
% 我们考虑提高NeRF在少视角的渲染效率问题.
% 虽然本文的方法没有在室内数据集上训练， 但是其实也能看到结合室内数据集先验的方法也有他们的优势，比如Dense Depth Priors对于平面重建的效果比我们的方法好。未来， 我们会考虑结合室内数据集丰富的先验，同时也能得到一个跨数据集泛化性能非常好的NeRF.

% Although our method achieve a better performance with sparse input views compared with previous methods, there are still many unsolved problems.

% if have a single appendix:
%\appendix[Proof of the Zonklar Equations]
% or
% \appendix  % for no appendix heading
% do not use \section anymore after \appendix, only \section*
% is possibly needed

% use appendices with more than one appendix
% then use \section to start each appendix
% you must declare a \section before using any
% \subsection or using \label (\appendices by itself
% starts a section numbered zero.)
%

% % use section* for acknowledgment
% \ifCLASSOPTIONcompsoc
%   % The Computer Society usually uses the plural form
%   \section*{Acknowledgments}
% \else
%   % regular IEEE prefers the singular form
%   \section*{Acknowledgment}
% \fi

% The authors would like to thank...

% Can use something like this to put references on a page
% by themselves when using endfloat and the captionsoff option.
\ifCLASSOPTIONcaptionsoff
  \newpage
\fi

% trigger a \newpage just before the given reference
% number - used to balance the columns on the last page
% adjust value as needed - may need to be readjusted if
% the document is modified later
%\IEEEtriggeratref{8}
% The "triggered" command can be changed if desired:
%\IEEEtriggercmd{\enlargethispage{-5in}}

% references section

% can use a bibliography generated by BibTeX as a .bbl file
% BibTeX documentation can be easily obtained at:
% http://mirror.ctan.org/biblio/bibtex/contrib/doc/
% The IEEEtran BibTeX style support page is at:
% http://www.michaelshell.org/tex/ieeetran/bibtex/
%\bibliographystyle{IEEEtran}
% argument is your BibTeX string definitions and bibliography database(s)
%\bibliography{IEEEabrv,../bib/paper}
%
% <OR> manually copy in the resultant .bbl file
% set second argument of \begin to the number of references
% (used to reserve space for the reference number labels box)
% \begin{thebibliography}{1}

% \bibitem{IEEEhowto:kopka}
% H.~Kopka and P.~W. Daly, \emph{A Guide to \LaTeX}, 3rd~ed.\hskip 1em plus
%   0.5em minus 0.4em\relax Harlow, England: Addison-Wesley, 1999.

% \end{thebibliography}

\appendices

\section{Additional results}
\begin{figure}[!h]
\centering
\includegraphics[width=1.0\linewidth]{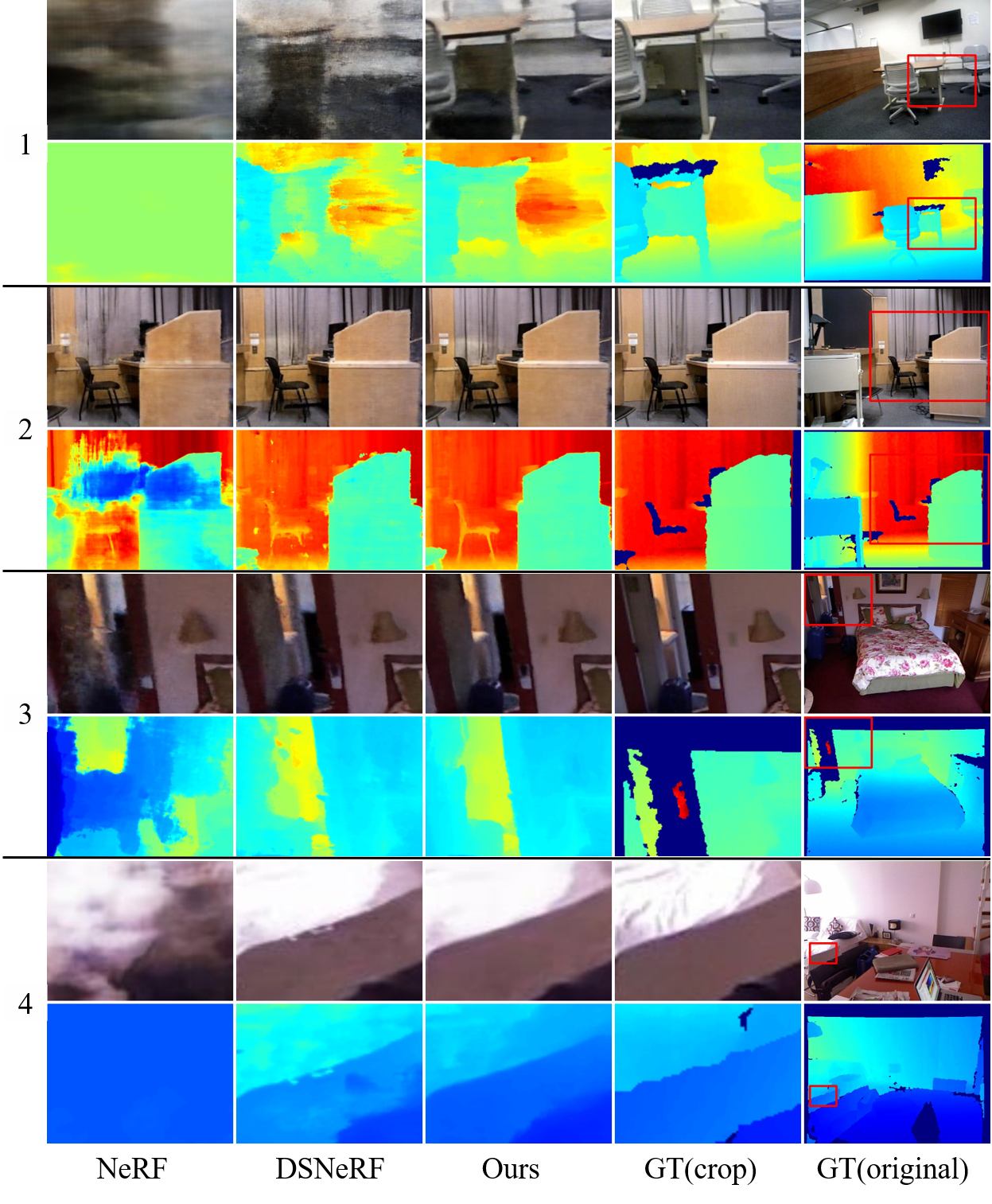}
\caption{Comparisons with method without pretraining on SUN3D.}
\label{fig:sun3d_cmp_without}
\end{figure}

\begin{figure}[!h]
\centering
\includegraphics[width=1.0\linewidth]{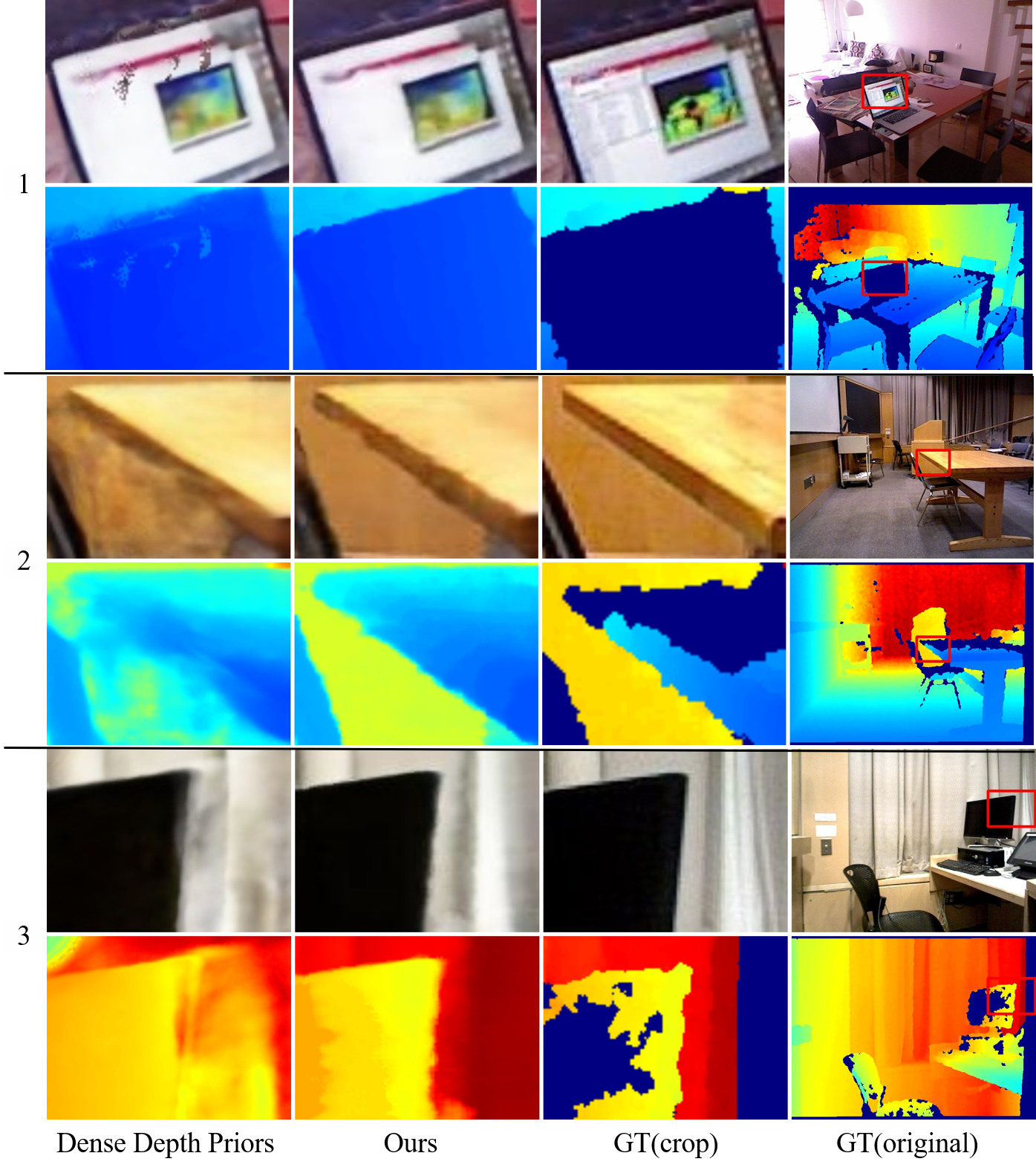}
\caption{Comparisons with method with pretraining on SUN3D.}
\label{fig:sun3d_cmp_with}
\end{figure}

More visualized results for comparisons on SUN3D can be found in Fig~\ref{fig:sun3d_cmp_without} and Fig~\ref{fig:sun3d_cmp_with}.

% \begin{figure*}[!h]
% \centering
% \includegraphics[width=1.0\linewidth]{fig/scannet_cmp_without.png}
% \caption{comparisons with method without pretraining on ScanNet.}
% \label{fig:scannet_cmp_without}
% \end{figure*}

% \begin{figure*}[!h]
% \centering
% \includegraphics[width=1.0\linewidth]{fig/scannet_cmp_with.png}
% \caption{comparisons with method with pretraining on ScanNet.}
% \label{fig:scannet_cmp_with}
% \end{figure*}

\subsection{Comparisons with NerfingMVS~\cite{nerfingmvs}}

% When applying the computation on entire rooms as opposed to a local region, the projected 3D points from other views frequently lie behind the camera. As a result the computed mean is often negative. Similarly, the computation of the near and far planes of the scenes is not suited for entire rooms, leading to a negative near plane in our case. Negative near plane and negative error map content lead to invalid sampling ranges, where the far bound lies in front of the near bound.

%在当应用For each input view an error map is computed by projecting the 3D points according to the depth prior to all other views, where a depth reprojection error is computed and normalized with the projected depth. The mean of the 4 smallest errors are used as values in the error map.

% 在NerfingMVS~\cite{nerfingmvs}中，对于每个输入视角，一个error map通过将3D points映射到所有的其他视角根据深度先验，然后一个深度的重投影误差计算出来，和被投影的深度一起做归一化，四个最小的errors的平均值作为error map中的值。
% 当应用这样的计算到整个房间时，与一个局部区域不同的是，来自其他视角的投影的3D点经常出现在相机后面。这样场景的near和far平面的计算不适合Room-scale scenes.负数的近平面导致了不合法的采样范围，远边界甚至出现在近边界的前面。

NerfingMVS~\cite{nerfingmvs} used an error map to guide NeRF sampling. For each input view, 3D points would be back-projected to the remaining views according to the depth prior. Then, NerfingMVS~\cite{nerfingmvs} calculated the depth re-projection error and defined error map as the mean of the top-4 smallest errors.

This procedure works fine in original NerfingMVS~\cite{nerfingmvs} data, which focus on local region depth estimation. However, when applying it to room-scale inputs, 3D points of projection from other views often falls behind the camera. The calculation of near and far planes of such scenes would result in negative values, and the far boundary even appears in front of the near boundary, leading to invalid sampling range~\cite{dense_depth_priors}.

\begin{figure}[!h]
\centering
\includegraphics[width=1.0\linewidth]{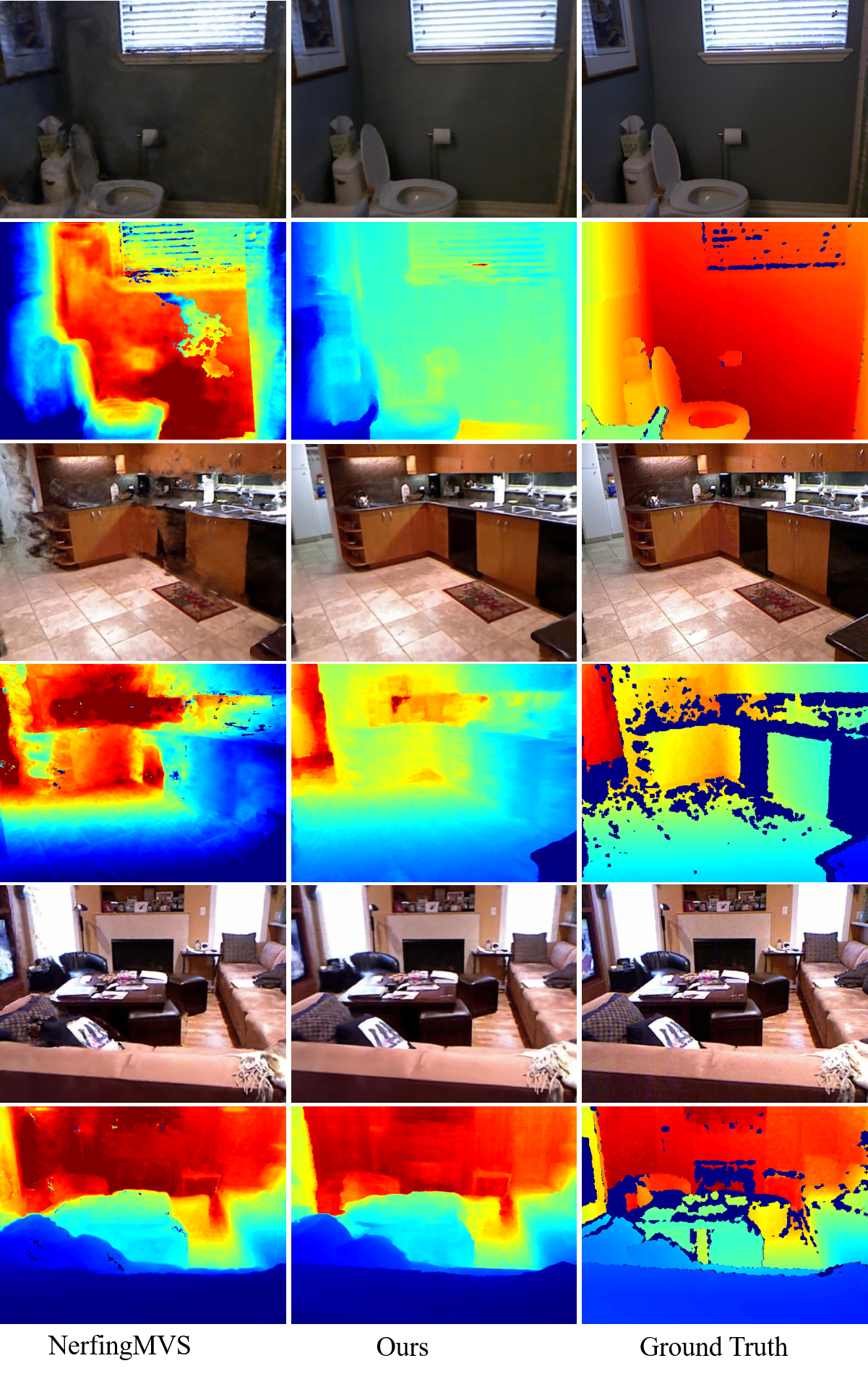}
\caption{Comparisons with NerfingMVS.}
\label{fig:nyu_cmp_nerfingmvs}
\end{figure}

%我们在跑nerfingmvs的官方代码发现，很多room-scale的场景无法跑，我们只能NYUv2上在少数场景上成功(nyuv2上的...)，在ScanNet和SUN3D上的8个样例数据集上都失败了。

We use the official code of NerfingMVS\footnote{\href{https://github.com/weiyithu/NerfingMVS}{https://github.com/weiyithu/NerfingMVS}} and find that it failed on all the scenes in ScanNet and Sun3D, only success on three scenes in NYUv2: bathroom\_0046, kitchen\_0026a and living\_room\_0033. The quantitative and qualitative comparisons on these scenes are shown in Tab.~\ref{tab_cmp_nerfingmvs} and Fig.~\ref{fig:nyu_cmp_nerfingmvs} respectively.

% 我们的方法在视角合成上的效果比NerfingMVS显著好，但是NerfingMVS重建平面的平坦度更好一些，因为它也利用了一个在室内数据集上预训练好的深度估计网络，这样的深度估计网络同样学习了很强的平面先验。
We can find that \sysname{} is significantly better than NerfingMVS in novel view synthesis and depth estimation. Only the predicted planes of NerfingMVS are slightly more flat than ours, but with the cost of an additional depth estimation network pretrained on a large amount of external indoor data. Such a network also learns strong plane priors like Dense Depth Priors~\cite{dense_depth_priors}.

%放一些对比图.

\begin{table*}[]
    \vspace{20pt}
    \centering
    \begin{tabular}{cccccccc}
        \hline
    {Method} & Need Pretraining?& PSNR$\uparrow$ &SSIM$\uparrow$&LPIPS$\downarrow$&Depth RMSE$\downarrow$ & Plane Mean Dev$\downarrow$  \\ \hline
NerfingMVS~\cite{nerfingmvs}  & Yes & 20.64 & 0.7348 & 0.2643 & 0.2190 & $\bm{0.0229}$ \\
Ours & No & $\bm{27.95}$ & $\bm{0.8569}$ & $\bm{0.1498}$ & $\bm{0.1960}$ & 0.0263 \\

        \hline       
    \end{tabular}
    \caption{Comparisons with NerfingMVS on NYUv2.}%Note: 我觉得第一列应该是指示帧编号的
\label{tab_cmp_nerfingmvs}
\end{table*}

% \section{Implementation Details}
% % Radiance Fields Our model architecture is based on NeRF [22]. The encoded position γ(x) is provided as input to the first of 8 layers as well as to the fifth, by concatenating it with the activations from the fourth layer. Layers 1–8 each have 256 neurons and ReLU activations. The output of layer 8 is passed through a single layer with softplus activation to produce density σ. The output of layer 8 is also passed through a 256-channel layer without activation, whose output is concatenated with the viewing direction d and the latent code `. The concatenated vector is fed to a 128-channel layer with ReLU activation, before the final layer producing the color c. The latent codes have a size of 4 on ScanNet and 16 on Matterport3D. For a fair ablation study, we chose a suitable depth loss weight λ for each approach and dataset, because the sparse depth input differs significantly on the two datsets (Tab. 4). The same λ is used across all scenes of each dataset.
% % \subsection{Radiance Fields}

\section{Datasets Details}
We take one frame every 20 frames from each video in the following datasets evenly for training and evaluation. 
 
\subsection{NYUv2}
The following eight scenes are used for evaluation:
\begin{itemize}
    \item bathroom\_0046
    \item dentist\_office\_0001
    \item dining\_room\_0004
    \item dining\_room\_0007
    \item dining\_room\_0016
    \item kitchen\_0026a
    \item living\_room\_0003
    \item living\_room\_0033
\end{itemize}

\subsection{ScanNet}
The following eight scenes are used for evaluation:
\begin{itemize}
    \item scene0419\_00
    \item scene0449\_00
    \item scene0477\_00
    \item scene0577\_00
    \item scene0753\_00
    \item scene0758\_00
    \item scene0776\_00
    \item scene0781\_00
\end{itemize}

\subsection{SUN3D}
The following eight scenes are used for evaluation:
\begin{itemize}
    \item home\_ac/home\_ac\_scan2\_2012\_aug\_22
    \item hotel\_stb/scan2
    \item mit\_3\_133/classroom\_3133\_nov\_6\_2012\_scan1\_erika
    \item mit\_gym\_z\_squash/gym\_z\_squash\_scan1\_oct\_26\_2012\_erika
    \item mit\_lab\_koch/lab\_koch\_bench\_nov\_2\_2012\_scan1\_erika
    \item mit\_w20\_athena/sc\_athena\_oct\_29\_2012\_scan1\_erika
    \item mit\_w85\_4/4\_2
    \item mit\_w85\_5/5\_1
    
\end{itemize}

% 0. 在SUN3D和ScanNet上的更多对比结果(可以放正文)
% 1. 网络参数(可以放正文)
% 2. 参数设定(可以放正文)
% 4. 其他方法的细节:比如权重*，
% 4. latent code实验
% 5. 数据集：对比的场景名称*

% 6.重建的可视化效果，定性对比(相对于DSNeRF和Dense Depth Priors的优势) 
% 3.加Latent code前后的实验结果（数据+图片）

\section{Latent Code}
\begin{table*}[]
    \vspace{20pt}
    \centering
    \begin{tabular}{lcccccccc}
        \hline
    Dataset & Method & Need Pretraining?& PSNR$\uparrow$ &SSIM$\uparrow$&LPIPS$\downarrow$&Depth RMSE$\downarrow$ & Plane Mean Dev$\downarrow$  \\ \hline
\multirow{2}{*}{NYUv2}&Ours  & No & 28.10 & 0.8561 & 0.1663 & 0.3113 & 0.0301 \\
&Ours+latent code  & No & 24.03(28.31) & 0.8255(0.8575) & 0.1824(0.1621) & 0.3136 & 0.0315 \\
        \hline   

\multirow{2}{*}{SUN3D}&Ours  & No & 22.88 & 0.7627 & 0.2748 & 0.5292 & 0.0327 \\
&Ours+latent code  & No & 18.90(24.65) & 0.7281(0.7914) & 0.2993(0.2516) & 0.4696 & 0.0318 \\
        \hline       

\multirow{2}{*}{Scannet}&Ours  & No & 24.67 & 0.8308 & 0.2481 & 0.2298 & 0.0270 \\
&Ours+latent code  & No & 23.73(25.92) & 0.8276(0.8420) & 0.2547(0.2340) & 0.2465 & 0.0264 \\
        \hline       

    \end{tabular}
    \caption{The quantitative results after adding the per-camera latent code to our method. Values in the parentheses are obtained by setting the latent code of the testing frames to the averaged latent code values of the two most adjacent frames.}%Note: 我觉得第一列应该是指示帧编号的
\label{tab_cmp_latent}
\end{table*}

% the latent codes used to represent view-specific appearance largely help to produce consistent colors across the scene. 

We also experimented on the per-frame latent code to deal with the illumination change across input images. According to NeRF-W~\cite{nerf-w}, latent codes can be used to control the view-specific appearance and is helpful for more consistent color prediction across the scene. Following NeRF-W, we set the dimension of latent code to 48 and concatenate it with view directions for input. Unfortunately, the latent codes in testing frames are never known and we set them to zero by default. The results after using the latent code are listed in Table~\ref{tab_cmp_latent}.

%其实，渲染图像的颜色并不用完全和评估的测试视角完全一样。为了弥补两者之间的差距，我们报告了一个额外的数值，这个数值是通过将测试视角最近的两个训练视角(采样中的编号相邻的两帧)的latent code的求平均值得到的。我们无法使用NeRF-W里左右划分评估步骤，在实际应用中也无法获取测试视角前后帧的训练视角，所以这个数值只是被当做性能上的上界。这些额外的数值被用括号列举出来了。
In fact, the colors of the rendered images do not have to be exactly the same as that of the test views. In order to compensate for the gap, we also report an additional set of PSNR, SSIM and LPIPS, which is obtained by setting the latent codes of the testing views to the averaged ones of the two nearest training views (two adjacent frames). However, in real-world applications, it is hard to know which two training frames are the most suitable for testing views and the left/right image split evaluation procedure in NeRF-W~\cite{nerf-w} is also impractical. Therefore, the reported values can only serve as the upper bound of performance (listed in parentheses). In the main text of our paper, we do not use the latent code for fair comparison. 

{\small
\bibliographystyle{plain}
\bibliography{ieee}
}

% biography section
% 
% If you have an EPS/PDF photo (graphicx package needed) extra braces are
% needed around the contents of the optional argument to biography to prevent
% the LaTeX parser from getting confused when it sees the complicated
% \includegraphics command within an optional argument. (You could create
% your own custom macro containing the \includegraphics command to make things
% simpler here.)

% or if you just want to reserve a space for a photo:
% \noindent\parbox{8.3cm}{\parpic{\includegraphics{fig/chenzheng.jpg}}{\small\quad {\bf Zheng Chen}  {received his B.S. degree in computer science and technology, Jilin University, Changchun, in 2020 and is currently a Ph.D. candidate of Tsinghua University, Beijing. His research interests include computer graphics and computer vision.} }\\[1mm]}

% % if you will not have a photo at all:
% \begin{IEEEbiographynophoto}{John Doe}
% Biography text here.
% \end{IEEEbiographynophoto}

% insert where needed to balance the two columns on the last page with
% biographies
%\newpage

% \begin{IEEEbiographynophoto}{Jane Doe}
% Biography text here.
% \end{IEEEbiographynophoto}

% You can push biographies down or up by placing
% a \vfill before or after them. The appropriate
% use of \vfill depends on what kind of text is
% on the last page and whether or not the columns
% are being equalized.

%\vfill

% Can be used to pull up biographies so that the bottom of the last one
% is flush with the other column.
%\enlargethispage{-5in}

% that's all folks
\end{document}